\documentclass{article}

\PassOptionsToPackage{numbers,compress,sort}{natbib}

 \usepackage[preprint]{neurips_2026}

\usepackage[table]{xcolor}
\definecolor{cvprblue}{rgb}{0.21,0.49,0.74}
\usepackage{colortbl}
\usepackage{pifont}
\usepackage{wrapfig}
\usepackage[utf8]{inputenc} % allow utf-8 input
\usepackage[T1]{fontenc}    % use 8-bit T1 fonts
\usepackage{hyperref}       % hyperlinks
\usepackage{url}            % simple URL typesetting
\usepackage{booktabs}       % professional-quality tables
\usepackage{amsfonts}       % blackboard math symbols
\usepackage{nicefrac}       % compact symbols for 1/2, etc.
\usepackage{microtype}      % microtypography
\usepackage{xcolor}         % colors
\usepackage{pgffor}
\usepackage{xstring}
\usepackage{hyperref}
\usepackage{graphicx}       % includegraphics
\newcommand{\eg}{\emph{e.g.}}
\newcommand{\ie}{\emph{i.e.}}
\newcommand{\paragrapht}[1]{\noindent\textbf{#1}}
\usepackage{amsmath}
\usepackage{multirow}
\usepackage[dvipsnames]{xcolor}
\definecolor{gradstart}{HTML}{F59E0B}  % 앰버
\definecolor{gradend}{HTML}{EF4444}    % 레드

\newcommand{\gradtext}[1]{%
    \StrLen{#1}[\strlen]%
    \foreach \i in {1,...,\strlen} {%
        \StrChar{#1}{\i}[\thechar]%
        \pgfmathtruncatemacro{\pct}{100 - (\i-1)*100/(\strlen-1)}%
        \textcolor{gradstart!\pct!gradend}{\thechar}%
    }%
}
\newcommand\blfootnote[1]{%
    \begingroup
    \renewcommand\thefootnote{}%
    \footnotetext{#1}%
    \endgroup
}
\newcommand{\ours}{TrackCraft3R}

\title{\ours: Repurposing Video Diffusion Transformers for Dense 3D Tracking}

\author{
    Jisu Nam\textsuperscript{\rm 1} \quad
    Jahyeok Koo\textsuperscript{\rm 1} \quad
    Soowon Son\textsuperscript{\rm 1} \quad
    Jaewoo Jung\textsuperscript{\rm 1} \quad \\
    {\bf Honggyu An\textsuperscript{\rm 1} \quad
    Junhwa Hur\textsuperscript{\rm 2}$^{\dagger}$ \quad
    Seungryong Kim\textsuperscript{\rm 1}$^{\dagger}$} \\[5pt]
    \textsuperscript{\rm 1}KAIST AI \qquad \textsuperscript{\rm 2}Google DeepMind \\[3pt]
{\tt \href{https://cvlab-kaist.github.io/TrackCraft3r}%
    {\gradtext{https://cvlab-kaist.github.io/TrackCraft3r}}}
}

\begin{document}

\maketitle

\begin{abstract}
Dense 3D tracking from monocular video is fundamental to dynamic scene understanding. While recent 3D foundation models provide reliable per-frame geometry, recovering object motion in this geometry remains challenging and benefits from strong motion priors learned from real-world videos. Existing 3D trackers either follow iterative paradigms trained from scratch on synthetic data or fine-tune 3D reconstruction models learned from static multi-view images, both lacking real-world motion priors. Pre-trained video diffusion transformers (video DiTs) offer rich spatio-temporal priors from internet-scale videos, making them a promising foundation for 3D tracking. However, their \emph{frame-anchored} formulation, which generates each frame's content, is fundamentally mismatched with \emph{reference-anchored} dense 3D tracking, which must follow the same physical points from a reference frame across time. We present \ours, the first method to repurpose a video DiT as a feed-forward dense 3D tracker. Given a monocular video and its frame-anchored reconstruction pointmap, \ours\ predicts a reference-anchored tracking pointmap that follows every pixel of the first frame across time in a single forward pass, along with its visibility. We achieve this through two designs: (i) a \emph{dual-latent representation} that uses per-frame geometry latents and reference-anchored track latents as dense queries, and (ii) \emph{temporal RoPE alignment}, which specifies the target timestamp of each track latent. Together, these designs convert the per-frame generative paradigm of video DiTs into a reference-anchored tracking formulation with LoRA fine-tuning. \ours\ achieves state-of-the-art performance on standard sparse and dense 3D tracking benchmarks, while running $1.3{\times}$ faster and using $4.6{\times}$ less peak memory than the strongest prior method. We further demonstrate robustness to large motions and long videos.
\end{abstract}

\blfootnote{$^\dagger$Co-corresponding.}

\section{Introduction}
\label{sec:introduction}
Recovering dense 3D trajectories from monocular video~\cite{feng2025st4rtrack,ngo2025deltav2,xiao2025spatialtrackerv2, wang2025scenetracker, cho2025seurat} is a fundamental building block for robotic manipulation~\cite{bharadhwaj2024track2act, kim2026pri4r}, dynamic scene reconstruction~\cite{liang2025zero, han2025d}, and controllable video generation~\cite{geng2025motion, nam2025emergent}. Because apparent motion is often dominated by camera ego-motion rather than object motion, accurate tracking requires reasoning in a 3D world coordinate frame in which camera motion is canceled out. Recent advances in monocular depth and pose estimation~\cite{li2025megasam, huang2025vipe, han2025emergent, lin2025depth} now provide reliable 3D geometry for arbitrary videos, enabling 3D trackers~\cite{xiao2025spatialtrackerv2, ngo2025deltav2, zhang2025tapip3d} to operate in a world coordinate frame where only residual object motion remains to be recovered.

Early 3D trackers~\cite{xiao2024spatialtracker, xiao2025spatialtrackerv2, wang2025scenetracker, ngo2024delta, ngo2025deltav2} follow the 2D tracker paradigm such as CoTracker paradigm~\cite{karaev2024cotracker,karaev2025cotracker3}, which iteratively updates trajectories based on local 3D correlation features, and is trained from scratch on synthetic 4D datasets~\cite{greff2022kubric, zheng2023pointodyssey, karaev2023dynamicstereo}. More recent feed-forward approaches~\cite{feng2025st4rtrack, karhade2025any4d, sucar2026v, liu2025trace} instead fine-tune pre-trained 3D reconstruction models~\cite{wang2024dust3r,leroy2024grounding,keetha2025mapanything}. While their pre-trained models offer strong spatial priors, they are learned from static multi-view images, lack rich temporal priors from real-world videos.

On the other hand, recent works demonstrate that pre-trained video diffusion models~\cite{blattmann2023stable, xing2024dynamicrafter}, especially video diffusion transformers (DiTs)~\cite{wan2025wan, kong2024hunyuanvideo, yang2024cogvideox}, already encode strong spatio-temporal priors from internet-scale real videos and effectively transfer to perception tasks such as video depth~\cite{zhang2026dvd, hu2025depthcrafter, shao2025learning}, camera pose~\cite{jiang2025geo4d}, and pointmap estimation~\cite{mai2025can}.

This motivates a key question: \textit{can we leverage the spatio-temporal priors of video DiTs for dense 3D tracking?} This is challenging because existing diffusion-based perception models produce \emph{frame-anchored} outputs (\ie, predictions defined independently at each frame~\cite{zhang2026dvd, hu2025depthcrafter, shao2025learning,jiang2025geo4d,mai2025can}), whereas dense 3D tracking requires \emph{reference-anchored} representations (\ie, tracking the same physical points from a reference frame across time). A concurrent work, MotionCrafter~\cite{zhu2026motioncrafter}, repurposes a video diffusion U-Net~\cite{blattmann2023stable} for 4D reconstruction, but predicts \emph{frame-anchored} scene flow between adjacent frames, requiring temporal chaining for dense 3D tracking and potentially leading to error accumulation, especially under occlusion.

In this paper, we introduce \textbf{\ours}, the first method that repurposes a video diffusion transformer~\cite{wan2025wan} as a feed-forward dense 3D tracker. Given a monocular video and its \emph{frame-anchored} reconstruction pointmap in world coordinates~\cite{huang2025vipe, lin2025depth, li2025megasam}, \ours\ predicts, in a single forward pass, a \emph{reference-anchored} tracking pointmap that tracks every pixel in the first frame across time, along with its visibility.

We achieve this by repurposing two core components of the video DiTs. First, we introduce a \textbf{dual-latent representation} consisting of (i) \emph{geometry latents}, which encode each frame’s RGB and reconstruction pointmap, and (ii) \emph{first-frame anchored track latents}, which encode the reference frame’s RGB and pointmap. The track latents act as dense query points defined in the first frame, while geometry latents represent 3D geometry over time in a shared world coordinate frame. Through full 3D attention, each track latent attends to geometry latents across frames to determine \emph{where} its corresponding point is and \emph{what} 3D position it should take. Second, we propose a \textbf{temporal RoPE alignment}, repurposing rotary positional embedding (RoPE)~\cite{su2024roformer} to encode the target timestamp of each track latent, specifying \emph{when} it attends to geometry latents. Together, \ours\ enables dense 3D tracking with LoRA~\cite{hu2022lora} fine-tuning, effectively converting the per-frame generative paradigm of video DiTs into a reference-anchored dense tracking paradigm. 

\ours\ achieves state-of-the-art performance on standard 3D sparse and dense tracking benchmarks~\cite{koppula2024tapvid, pan2023aria, joo2015panoptic, karaev2023dynamicstereo, zheng2023pointodyssey, greff2022kubric}. Notably, \ours\ runs $1.3{\times}$ faster 
and uses $4.6{\times}$ less peak memory than the state-of-the-art 3D tracker DELTAv2~\cite{ngo2025deltav2}. We further demonstrate robustness to large motions and long videos, and extensive ablations validate our design choices.

In summary, our contributions are threefold:
(1) we present \textbf{\ours}, the first method to repurpose a video diffusion transformer for feed-forward dense 3D tracking;
(2) we propose a dual-latent representation and temporal RoPE alignment to convert frame-anchored generation into first-frame-anchored dense 3D tracking; and
(3) we achieve state-of-the-art performance on standard 3D tracking benchmarks, while demonstrating robustness to large temporal strides and long videos.

\section{Related Work}

\paragrapht{3D Point Tracking.}
Point tracking aims to recover long-range motion trajectories in videos. Early 2D tracking methods~\cite{sand2008particle, doersch2022tap, harley2022particle, doersch2023tapir, karaev2024cotracker, karaev2025cotracker3, cho2024local} iteratively refine trajectories within sliding temporal windows. To extend this to 3D, several works incorporate monocular depth~\cite{yang2024depth, lin2025depth} and track in camera coordinates~\cite{xiao2024spatialtracker, wang2025scenetracker, ngo2024delta, ngo2025deltav2}, while others~\cite{xiao2025spatialtrackerv2, zhang2025tapip3d} further utilize camera poses~\cite{huang2025vipe, lin2025depth, li2025megasam} to operate in a world coordinate frame, where camera motion is explicitly compensated. However, these methods rely on iterative trajectory updates and are trained from scratch on synthetic 4D datasets~\cite{greff2022kubric, zheng2023pointodyssey, karaev2023dynamicstereo}.

Recent feed-forward approaches~\cite{feng2025st4rtrack, liang2025zero, han2025d, jin2024stereo4d, liu2025trace, karhade2025any4d, sucar2026v} instead propose to fine-tune pre-trained 3D reconstruction models~\cite{wang2024dust3r, leroy2024grounding, keetha2025mapanything, yang2025fast3r} on synthetic 4D data. 
While these methods benefit from strong spatial priors of pre-trained models, they still lack strong temporal priors from real-world video dynamics. A concurrent work, MotionCrafter~\cite{zhu2026motioncrafter}, incorporates temporal priors by repurposing a video diffusion U-Net~\cite{blattmann2023stable} for 4D reconstruction. However, it predicts frame-anchored scene flow between adjacent frames, requiring temporal chaining that accumulates errors under occlusion. In contrast, \ours\ repurposes a video diffusion transformer to directly produce reference-anchored tracking pointmap in a single forward pass, avoiding temporal chaining.

\paragrapht{Video Diffusion Models for Frame-Anchored Perception.} Image diffusion models have been successfully repurposed for a wide range of perception tasks, including depth estimation~\cite{ke2024repurposing, he2024lotus}, surface normal prediction~\cite{fu2024geowizard, he2024lotus}, dense correspondence~\cite{nam2023diffusion, tang2023emergent, hedlin2023unsupervised}, and optical flow~\cite{saxena2023surprising}. This paradigm has naturally extended to the video domain, where video diffusion models provide robust spatio-temporal priors. Early works repurpose video diffusion U-Nets~\cite{blattmann2023stable, xing2024dynamicrafter} for temporally consistent video depth estimation~\cite{hu2025depthcrafter, shao2025learning}, per-frame pointmap estimation~\cite{xu2025geometrycrafter}, and joint estimation of depth, pointmaps, and ray maps~\cite{jiang2025geo4d}.
Recently video diffusion transformers (DiTs)~\cite{wan2025wan, kong2024hunyuanvideo, yang2024cogvideox} has driven performance improvement across multiple tasks: DVD~\cite{zhang2026dvd} repurposes the Wan 2.1 DiT~\cite{wan2025wan} for video depth, and Sora3R~\cite{mai2025can} adapts an OpenSora DiT for pointmap prediction. 

Despite the diversity of tasks, all these methods produce {frame-anchored} outputs, where predictions are tied to the content and timestamp of individual frames. Dense 3D tracking, by contrast, requires {reference-anchored} predictions that follow the same physical content from a reference frame across time. To the best of our knowledge, \ours\ is the first to repurpose a video DiT for {reference-anchored} dense 3D tracking. A recent work~\cite{son2025repurposing} leverages video DiT features for sparse 2D point tracking. However, this method adds a tracking head (\eg, a CoTracker head~\cite{karaev2025cotracker3}) on top of the video DiT features, rather than repurposing the video DiT itself.

\section{Preliminaries}
\label{sec:preliminary}

\paragrapht{Variational Autoencoder (VAE).}
A VAE encoder $\mathcal{E}$ maps a video $\mathbf{V} \in \mathbb{R}^{(1+F) \times H \times W \times 3}$ into a latent representation $\mathbf{z} = \mathcal{E}(\mathbf{V}) \in \mathbb{R}^{(1+f) \times h \times w \times c}$, where $H$, $W$, and $(1{+}F)$ denote the spatial resolution and number of frames, and $h$, $w$, and $(1{+}f)$ denote their spatially and temporally downsampled counterparts. $c$ is the latent channel dimension. Here, temporal downsampling is applied only to the $F$ frames, while the first frame is preserved. A decoder $\mathcal{D}$ reconstructs the video from $\mathbf{z}$.

Prior works show that VAEs pre-trained on RGB videos can be repurposed to encode and decode geometric modalities such as pointmaps~\cite{mai2025can, xu2025geometrycrafter}, depth maps~\cite{hu2025depthcrafter, zhang2026dvd}, and camera rays~\cite{jiang2025geo4d, jang2026rays}, enabling diffusion models to operate in this latent space for geometric prediction.

\paragrapht{Video Diffusion Transformers (DiTs).}
The latent $\mathbf{z}$ is patchified and projected, and a transformer $f_\theta$ is trained with rectified flow matching~\cite{lipman2022flow} to predict the velocity field along a linear interpolation between noise and data. The model applies full 3D attention, where each token $i$ produces query $\mathbf{q}_i$, key $\mathbf{k}_i$, and value $\mathbf{v}_i$, and attends to all the other tokens $j$ with weights proportional to $\mathbf{q}_i^\top \mathbf{k}_j / \sqrt{d_k}$, where $d_k$ is the key dimension. 

In this work, following~\cite{he2024lotus, zhang2026dvd}, we repurpose $f_\theta$ as a feed-forward regressor rather than a multi-step denoiser, enabling efficient inference without iterative sampling.

\paragrapht{3D Rotary Positional Embedding (3D RoPE).}
To encode relative spatio-temporal structure, video DiTs employ 3D RoPE~\cite{su2024roformer}. The channel dimension of each query and key vector is partitioned into temporal and spatial groups, and axis-specific rotation matrices are applied on each token's 3D position $\mathbf{p}_i = (x_i, y_i, t_i)$, where $(x_i, y_i)$ denote spatial coordinates and $t_i$ denotes the temporal index. Under RoPE, the attention score between tokens $i$ and $j$ becomes 
\begin{equation}
\tilde{\mathbf{q}}_i^\top \tilde{\mathbf{k}}_j = \mathbf{q}_i^\top \mathbf{R}_{\mathbf{p}_j - \mathbf{p}_i} \mathbf{k}_j,
\label{eq:rope}
\end{equation} where $\tilde{\mathbf{q}}_i$ and $\tilde{\mathbf{k}}_j$ denote the query and key vectors after applying RoPE. $\mathbf{R}_{\mathbf{p}_j - \mathbf{p}_i}$ is a block-diagonal rotation matrix parameterized by the relative offset $\mathbf{p}_j - \mathbf{p}_i$. Thus, attention depends only on relative positions, \ie,~tokens with similar $t_i$ interact more strongly.

\begin{figure*}[t]
  \centering
  \includegraphics[width=\textwidth]{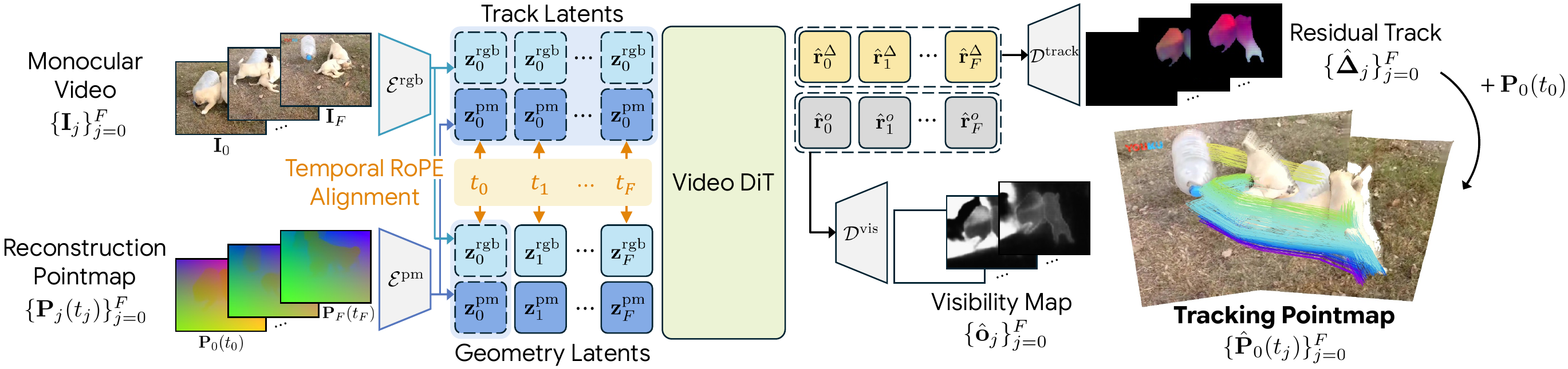}
\caption{\textbf{Overall architecture.}
Each RGB frame $\mathbf{I}_j$ and its reconstruction pointmap $\mathbf{P}_j(t_j)$ are encoded into RGB and pointmap latents using separate VAE encoders. A geometry latent is formed by channel-wise concatenation, and a track latent replicates the first-frame geometry latent across all frames. The latents are concatenated along the token dimension and processed by a video DiT, where RoPE assigns the same temporal index to each frame. The track latent outputs are decoded using separate VAE decoders into a residual track $\hat{\mathbf{\Delta}}_j$ and visibility $\hat{\mathbf{o}}_j$.}
  \label{fig:architecture}\vspace{-10pt}
\end{figure*}
\section{Video Diffusion Transformer for Dense 3D Tracking} 

We present a novel framework that densely tracks dynamic video content in a 3D world coordinate frame in a single forward pass. Recent 3D foundation models for depth and camera pose~\cite{huang2025vipe, lin2025depth, li2025megasam} provide reliable 3D scene geometry in world coordinates for arbitrary videos. Building on the pre-trained spatio-temporal priors of video diffusion transformers (DiTs), we leverage this 3D geometry as input and repurpose a video DiT to regress dense 3D tracks directly in this coordinate frame.

Specifically, we adopt two pointmap representations~\cite{feng2025st4rtrack, sucar2026v, han2025d} that encode 3D geometry and motion: a \emph{frame-anchored} pointmap as input and a \emph{reference-anchored} pointmap as output. In Sec.~\ref{sec:formulation}, we formulate these pointmaps and define the problem.

However, the frame-anchored generative paradigm of video DiTs is fundamentally misaligned with dense 3D tracking, which requires reference-anchored predictions of the same physical points across time. To address this, we repurpose a video DiT with \emph{dual-latent representation} and \emph{temporal RoPE alignment}. Sec.~\ref{sec:architecture} provides further details on the model architecture.

\begin{wrapfigure}[19]{r}{0.49\textwidth}
  \centering
  \vspace{-15pt}
  \includegraphics[width=\linewidth]{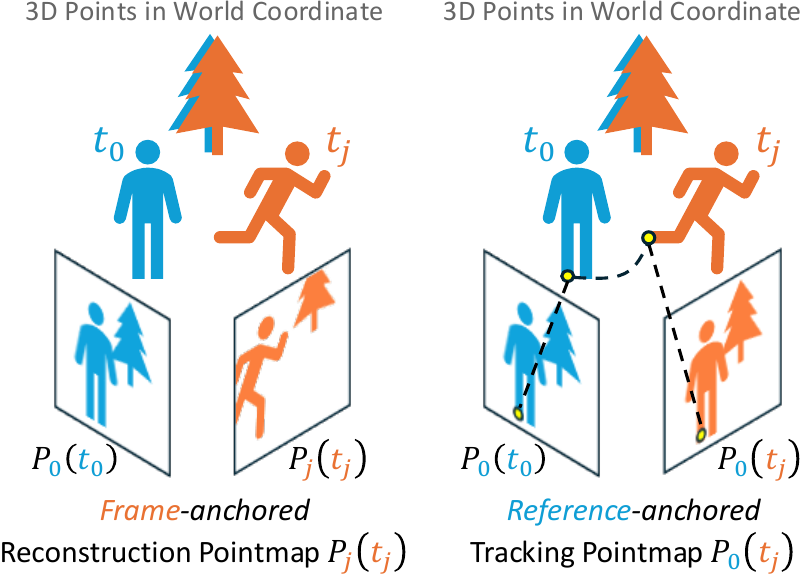}
  \vspace{-15pt}
  \caption{\textbf{Pointmap Representations.}
Given 3D points of a dynamic scene in world coordinates, the reconstruction pointmap represents 3D points of $\mathbf{I}_j$'s content at $t_j$, while the tracking pointmap represents 3D points of $\mathbf{I}_0$ (reference frame) at $t_j$, so that all 3D points of $\mathbf{I}_0$ are tracked across time.}
  \label{fig:pointmap}
\end{wrapfigure}

\subsection{Problem Formulation}
\label{sec:formulation}

Following~\cite{feng2025st4rtrack, sucar2026v}, given a monocular video $\mathbf{V} = \{\mathbf{I}_t\}_{t=0}^{F} \in \mathbb{R}^{(1+F) \times H \times W \times 3}$, we define a time-dependent pointmap $\mathbf{P}_{{i}}({t_j}) \in \mathbb{R}^{H \times W \times 3}$ 
as the 3D positions of the {physical content observed in frame $\mathbf{I}_i$} at {timestamp $t_j$}. This provides a unified representation of dynamic scenes, jointly encoding 3D geometry and motion. 
All pointmaps are expressed in a shared world coordinate frame (we use the first frame as the reference frame), and we omit the coordinate index for simplicity. 

\paragrapht{Reconstruction Pointmap.}
Each frame $\mathbf{I}_j \in \mathbb{R}^{H \times W \times 3}$ is lifted to 3D using depth and camera intrinsics, and transformed into the shared world coordinate frame via camera extrinsics. This yields a \emph{frame-anchored} reconstruction pointmap 
$\mathbf{P}_{\textcolor{RedOrange}{j}}(\textcolor{RedOrange}{t_j}) \in \mathbb{R}^{H \times W \times 3}$, 
which represents the 3D positions of the content in frame \textcolor{RedOrange}{$\mathbf{I}_j$} at its own timestamp \textcolor{RedOrange}{$t_j$}. Note that such pointmaps can be readily obtained either from ground-truth~\cite{zheng2023pointodyssey, karaev2023dynamicstereo, greff2022kubric} or from estimated depth and camera pose using recent 3D foundation models~\cite{huang2025vipe, lin2025depth, li2025megasam}.

\paragrapht{Tracking Pointmap.}
To enable tracking, we define a \emph{reference-anchored} tracking pointmap 
$\mathbf{P}_{\textcolor{NavyBlue}{0}}(\textcolor{RedOrange}{t_j}) \in \mathbb{R}^{H \times W \times 3}$, 
which represents the 3D positions of the content originally observed in the reference frame \textcolor{NavyBlue}{$\mathbf{I}_0$} 
at timestamp \textcolor{RedOrange}{$t_j$}. 
Here, the reference index is fixed to $0$ while time varies, so the same physical points from $\mathbf{I}_0$ are tracked consistently across frames. Fig.~\ref{fig:pointmap} illustrates both pointmaps.

\paragrapht{Our Objective.}
Given a video $\mathbf{V} = \{\mathbf{I}_j\}_{j=0}^{F}$ and its reconstruction pointmaps
$\{\mathbf{P}_j(t_j)\}_{j=0}^{F}$, which provide per-frame 3D geometry in a shared world coordinate frame, 
our goal is to predict the tracking pointmaps 
$\{\mathbf{P}_0(t_j)\}_{j=0}^{F}$ that establish dense 3D correspondences across time by tracking the physical content of the reference frame $\mathbf{I}_0$ throughout the sequence. 
In addition, we predict visibility maps 
$\{\mathbf{o}_j\}_{j=0}^{F}$, where $\mathbf{o}_j \in [0,1]^{H \times W}$ indicates whether each tracked point from $\mathbf{I}_0$ is visible at time $t_j$.

\subsection{Model Architecture}
\label{sec:architecture}

An overview of our architecture is shown in Fig.~\ref{fig:architecture}.  
Given a video $\{\mathbf{I}_j\}_{j=0}^{F}$ and its reconstruction pointmaps $\{\mathbf{P}_j(t_j)\}_{j=0}^{F}$, we encode each RGB frame and pointmap independently using separate VAE encoders $\mathcal{E}^{\text{rgb}}$ and $\mathcal{E}^{\text{pm}}$, yielding per-frame RGB latents $\mathbf{z}_j^{\text{rgb}}$ and pointmap latents $\mathbf{z}_j^{\text{pm}}$:
\begin{equation}
\mathbf{z}_j^{\text{rgb}} = \mathcal{E}^{\text{rgb}}(\mathbf{I}_j) \in \mathbb{R}^{h \times w \times c}, 
\quad
\mathbf{z}_j^{\text{pm}} = \mathcal{E}^{\text{pm}}(\mathbf{P}_j(t_j)) \in \mathbb{R}^{h \times w \times c}.
\end{equation}

To preserve per-frame spatial precision, we bypass temporal compression in the original 3D VAE by treating the temporal dimension as a batch dimension~\cite{nam2025emergent} (see the ablation in Tab.~\ref{tab:ablation_design}).

\paragrapht{Point Map Normalization.}
Prior to VAE encoding, each pointmap is normalized by subtracting the mean and dividing by the maximum distance from the mean, both computed over points whose depths fall within the 2\%–98\% percentile range across all frames to exclude outliers. As a result, the normalized values lie approximately within $[-1, 1]$.

\paragrapht{Dual-Latent Representation.}
To repurpose a video DiT for reference-anchored 3D tracking, we define two types of latents for the model input: a \emph{geometry latent} $\mathbf{g}_j$, which encodes 3D geometry at timestamp $t_j$, and a \emph{first-frame-anchored track latent} $\mathbf{r}_j$, which serves as a dense query anchored to the reference frame $\mathbf{I}_0$ for tracking across time.

To explicitly couple RGB appearance $\mathbf{z}_j^{\text{rgb}}$ and 3D geometry $\mathbf{z}_j^{\text{pm}}$ at each spatial location, the geometry latent $\mathbf{g}_j$ is formed by channel-wise concatenation at timestamp $t_j$. To anchor tracking to the reference frame, the track latent $\mathbf{r}_j$ is obtained by replicating the first-frame geometry latent across all timestamps:
\begin{equation}
\mathbf{g}_j = [\mathbf{z}_j^{\text{rgb}};\; \mathbf{z}_j^{\text{pm}}] \in \mathbb{R}^{h \times w \times 2c}, 
\qquad
\mathbf{r}_j = \mathbf{g}_0 \in \mathbb{R}^{h \times w \times 2c},
\end{equation}
where $[\cdot;\cdot]$ denotes channel-wise concatenation.

We concatenate the geometry and track latents along the token dimension and process them with a video DiT $f_\theta$:
\begin{equation}
\label{eq:dit_output}
\{\hat{\mathbf{r}}_j\}_{j=0}^{F} = f_\theta\big([\{\mathbf{g}_j\}_{j=0}^{F},\; \{\mathbf{r}_j\}_{j=0}^{F}]\big),
\end{equation}
where $[\cdot,\cdot]$ denotes concatenation along the token sequence dimension. 
The outputs corresponding to the track latents, $\hat{\mathbf{r}}_j \in \mathbb{R}^{h \times w \times 2c}$, are used for tracking pointmap and visibility prediction. 

Intuitively, RGB latents provide cues for spatial matching, while pointmap latents store the associated 3D positions. Once $\mathbf{z}_0^{\text{rgb}}(u_r, v_r)$ in the track latent $\mathbf{r}_j$ matches the same physical point as $\mathbf{z}_j^{\text{rgb}}(u_g, v_g)$ in the geometry latent $\mathbf{g}_j$ via attention, the corresponding pointmap latent $\mathbf{z}_j^{\text{pm}}(u_g, v_g)$ directly provides its 3D position $\mathbf{P}_j(t_j)(u_g, v_g)$, which defines the tracked point $\mathbf{P}_0(t_j)(u_r, v_r)$. Here, $(u_r, v_r)$ denotes spatial coordinates in the track latent, and $(u_g, v_g)$ denotes the corresponding spatial coordinates in the geometry latent.

To convert the video DiT into a one-step regressor, we fix the diffusion timestep to zero and use a null text prompt. We further evaluate the inference efficiency of our one-step model in Tab.~\ref{tab:efficiency}.

% Two-row attention visualization figure (all indices 0-indexed).
% Row (a): query overlay (r_5) + 3 key-frame attention maps (g_2, g_5, g_8) at layer 15
% Row (b): query overlay (r_5) + 3 layer-specific attention maps (layers 14, 15, 16) at g_5
%
% Image aspect: 832 x 480 → height = width * 480/832 ≈ 0.577 * width
%   width=0.24\textwidth → height ≈ 0.1385\textwidth
%
% PDFs needed (place in figures/ alongside this .tex):
%   Row (a): figA_query.pdf, figA_g02_layer15.pdf, figA_g05_layer15.pdf, figA_g08_layer15.pdf
%   Row (b): figB_query.pdf, figB_layer14.pdf, figB_layer15.pdf, figB_layer16.pdf
%
% Usage in main paper:  \input{figures/attention_vis}

\begin{figure*}[t]
  \centering
  \setlength{\tabcolsep}{1pt}
  \renewcommand{\arraystretch}{0.4}

  % ---------- Row (a): r_5 → g_j across {g_2, g_5, g_8} ----------
  \noindent
  \begin{minipage}[c]{0.010\textwidth}
    \centering\footnotesize (a)
  \end{minipage}\hspace{3pt}%
  \begin{minipage}[c]{0.97\textwidth}
    \centering
    \begin{tabular}{@{}c@{\hspace{1pt}}c@{\hspace{1pt}}c@{\hspace{1pt}}c@{}}
      \includegraphics[width=0.24\textwidth,height=0.1385\textwidth]{figures/figA_query} &
      \includegraphics[width=0.24\textwidth,height=0.1385\textwidth]{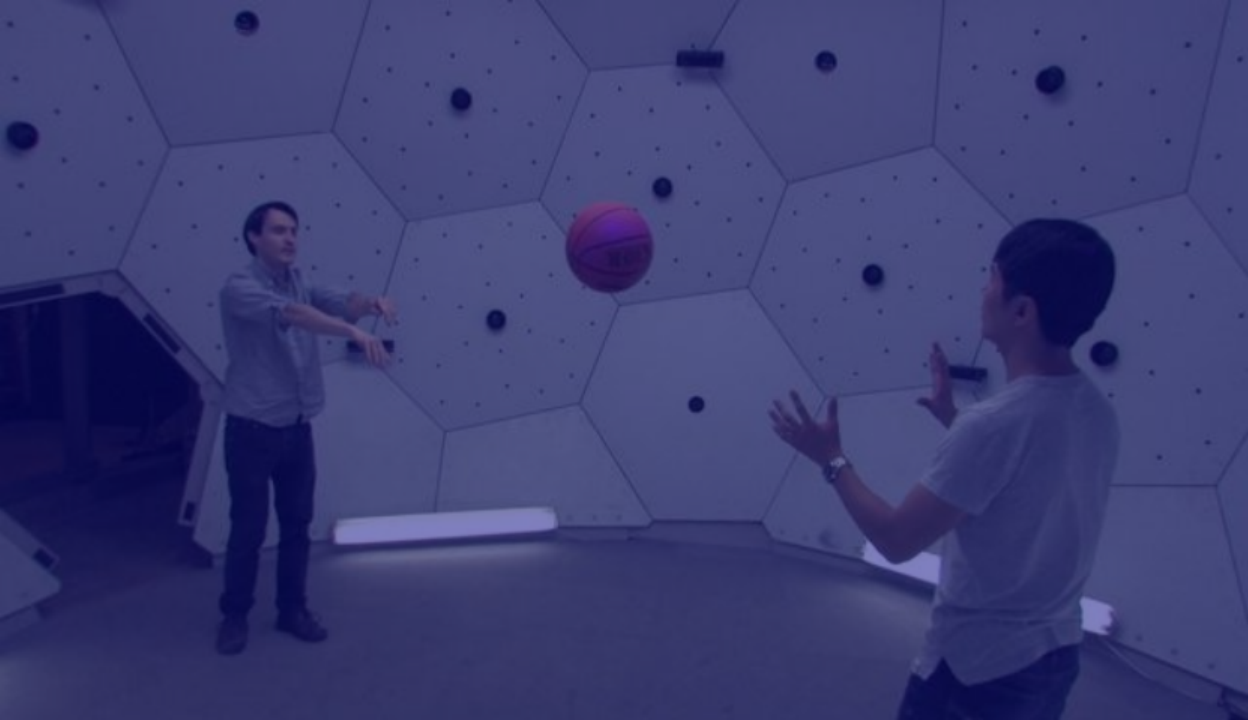} &
      \includegraphics[width=0.24\textwidth,height=0.1385\textwidth]{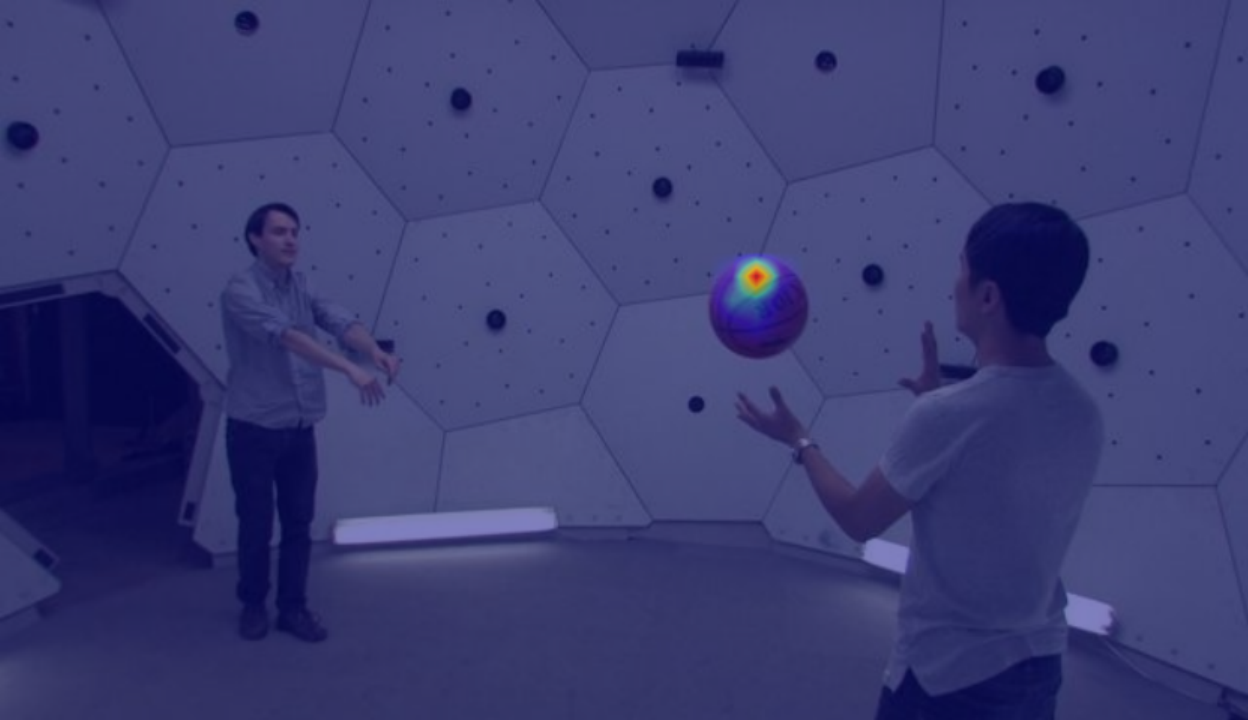} &
      \includegraphics[width=0.24\textwidth,height=0.1385\textwidth]{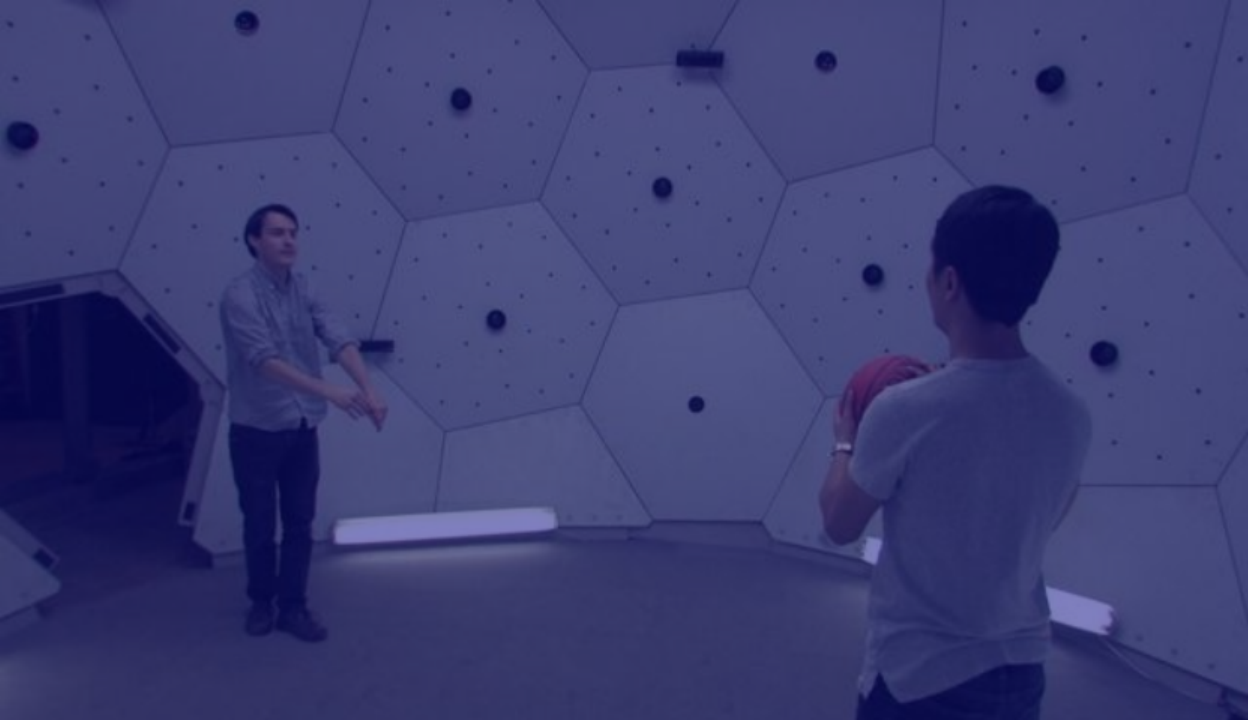} \\[2pt]
      \footnotesize $\mathbf{r}_5$, Query point &
      \footnotesize $\mathbf{g}_2$ &
      \footnotesize $\mathbf{g}_5$ &
      \footnotesize $\mathbf{g}_8$ \\
    \end{tabular}
  \end{minipage}

  \vspace{1pt}

  % ---------- Row (b): r_5 → g_5 across selected layers ----------
  \noindent
  \begin{minipage}[c]{0.010\textwidth}
    \centering\footnotesize (b)
  \end{minipage}\hspace{3pt}%
  \begin{minipage}[c]{0.97\textwidth}
    \centering
    \begin{tabular}{@{}c@{\hspace{1pt}}c@{\hspace{1pt}}c@{\hspace{1pt}}c@{}}
      \includegraphics[width=0.24\textwidth,height=0.1385\textwidth]{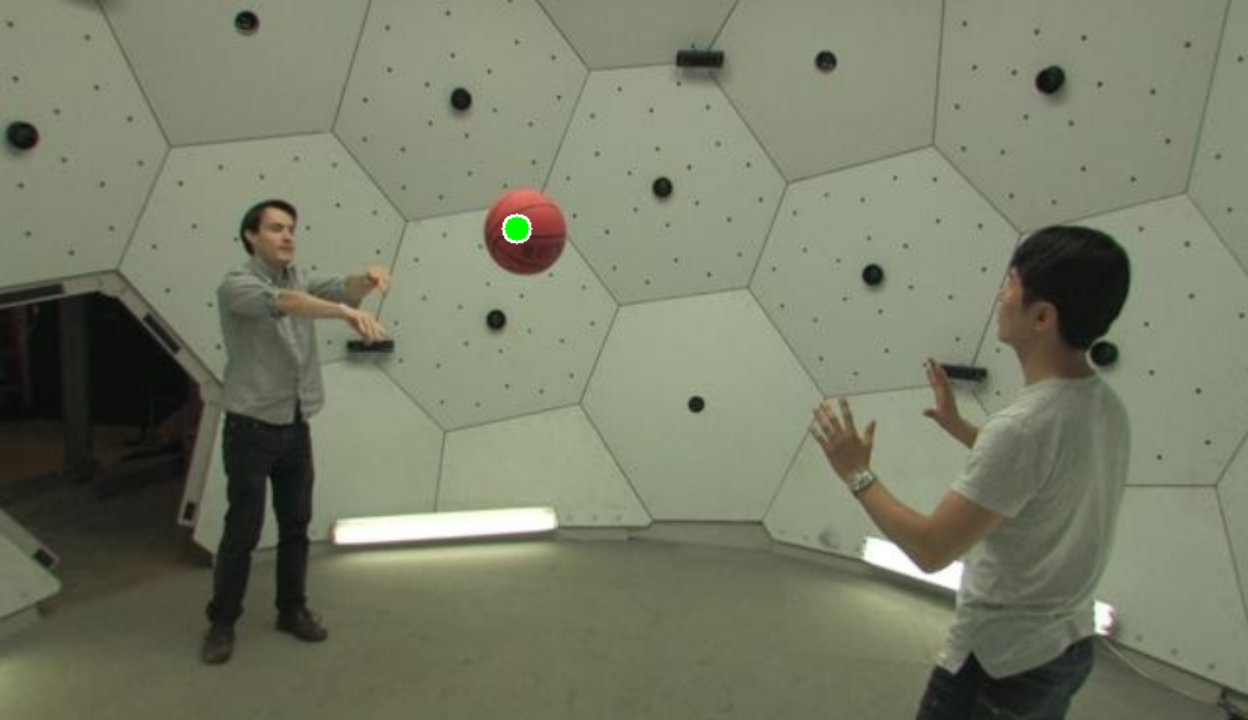} &
      \includegraphics[width=0.24\textwidth,height=0.1385\textwidth]{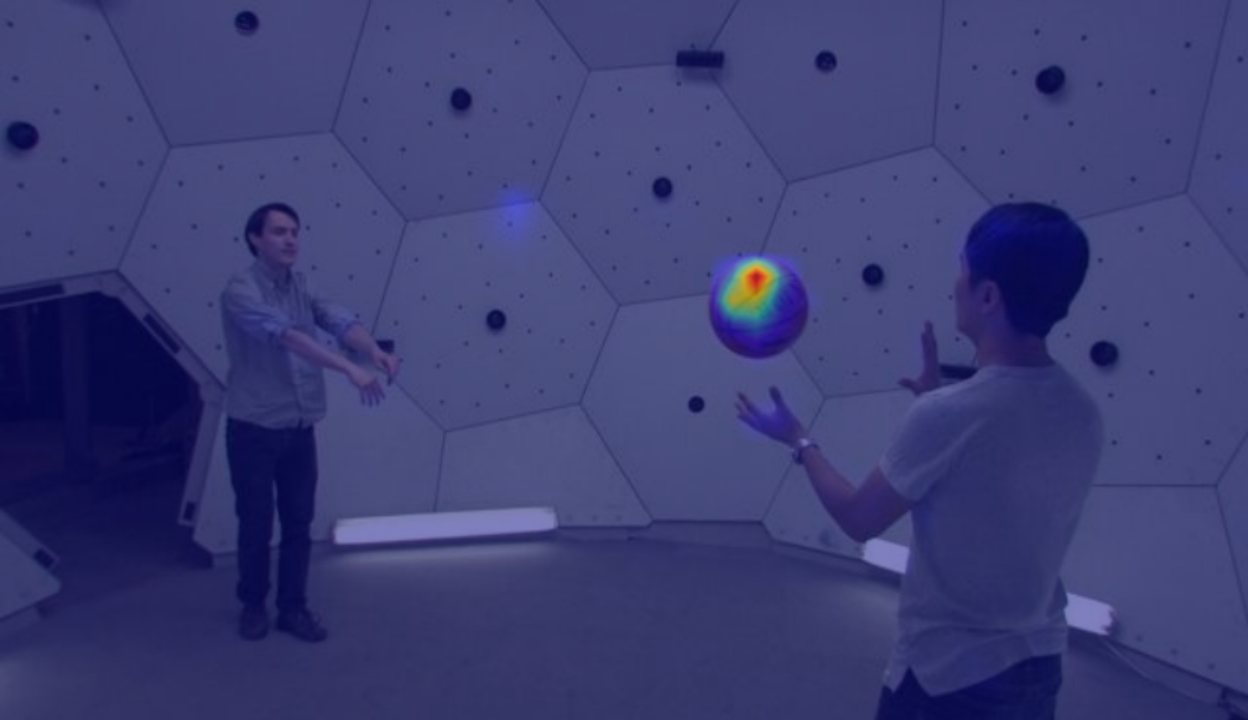} &
      \includegraphics[width=0.24\textwidth,height=0.1385\textwidth]{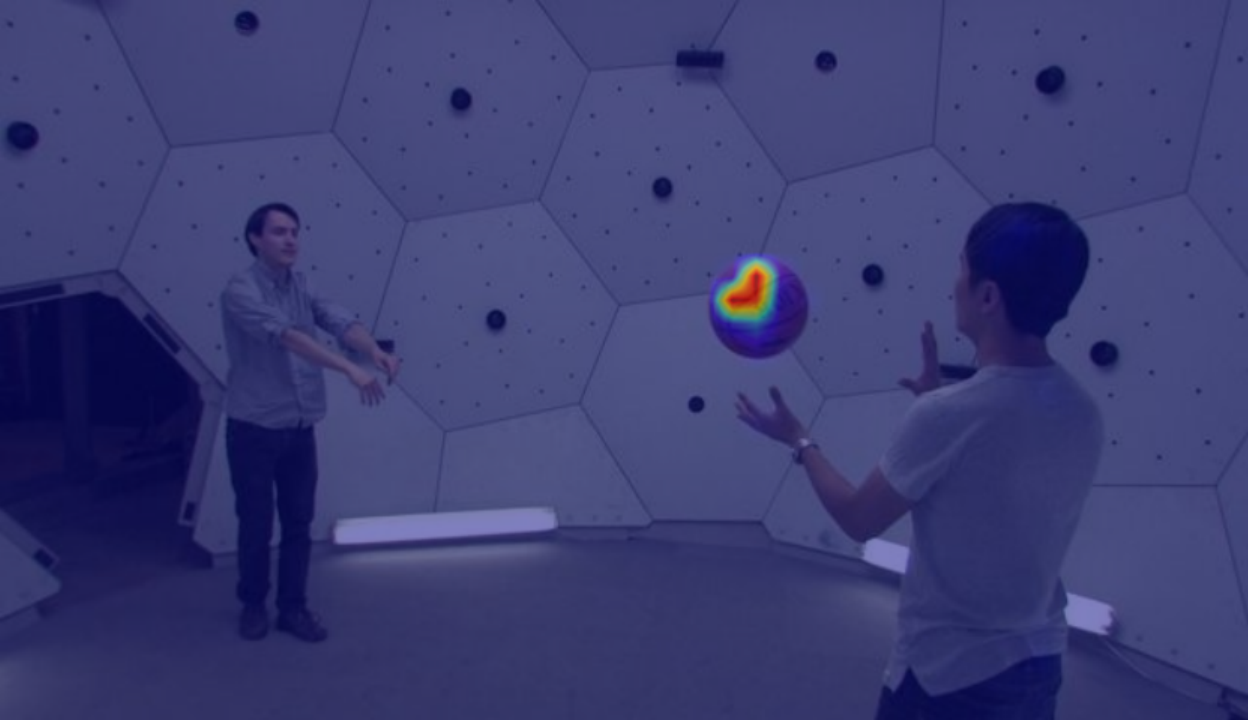} &
      \includegraphics[width=0.24\textwidth,height=0.1385\textwidth]{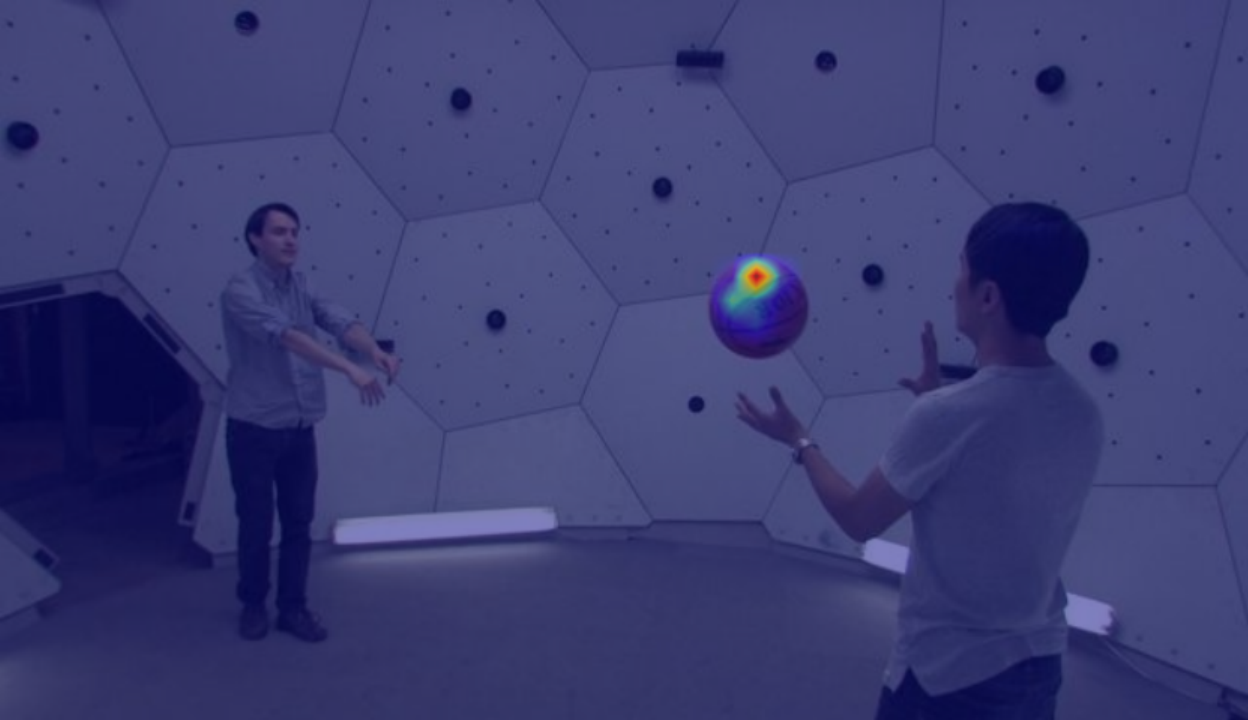} \\[2pt]
      \footnotesize $\mathbf{r}_5$, Query point &
      \footnotesize $\mathbf{g}_5$, Layer 14 &
      \footnotesize $\mathbf{g}_5$, Layer 15 &
      \footnotesize $\mathbf{g}_5$, Layer 16 \\
    \end{tabular}
  \end{minipage}

  \caption{\textbf{Query-key attention visualization.} The query point is marked as a {green circle}. (a) Attention from track latents $\mathbf{r}_5$ at timestamp $t_5$ to geometry latents $\{\mathbf{g}_j\}_{j=0}^{F}$. Attention is predominantly localized on $\mathbf{g}_5$, showing that RoPE correctly assigns a target timestamp to each track latent.
  (b) Within $\mathbf{g}_5$, attention aligns with the same physical point under motion, demonstrating accurate dense correspondence between track and geometry latents.}
  \label{fig:attention}\vspace{-10pt}
\end{figure*}

\paragrapht{Temporal RoPE Alignment.}
To ensure that each track latent attends to the geometry latent at the correct timestamp, we utilize the temporal axis of 3D RoPE~\cite{su2024roformer}. As illustrated in Fig.~\ref{fig:architecture}, we assign both $\mathbf{g}_j$ and $\mathbf{r}_j$ the same temporal RoPE index $t_j$ (Eq.~\ref{eq:rope}). Since RoPE encodes relative position, tokens with identical temporal indices exhibit stronger attention. Consequently, each track latent $\mathbf{r}_j$ attends to the geometry latent $\mathbf{g}_j$ at timestamp $t_j$, retrieving the corresponding 3D position.

Fig.~\ref{fig:attention}(a) visualizes the query–key attention from $\mathbf{r}_5$ to $\{\mathbf{g}_k\}_{k=0}^{F}$, showing that attention is predominantly localized on $\mathbf{g}_5$, confirming that temporal RoPE alignment correctly specifies the target timestamp. Fig.~\ref{fig:attention}(b) further visualizes the attention between $\mathbf{r}_5$ and $\mathbf{g}_5$ across different transformer layers, showing that full 3D attention effectively establishes accurate correspondences between track and geometry latents under motion. Full attention visualizations and additional discussion are provided in the Appendix~\ref{sec:appendix_attention}.

\paragrapht{Trajectory and Visibility Prediction.}
We decode the video DiT outputs corresponding to the track latents, $\hat{\mathbf{r}}_j$, into a tracking pointmap $\hat{\mathbf{P}}_0(t_j)$ and a visibility map $\hat{\mathbf{o}}_j$.
The latent $\hat{\mathbf{r}}_j \in \mathbb{R}^{h \times w \times 2c}$ is channel-wise partitioned into two components: the first half is used for pointmap prediction, and the second half for visibility prediction.

Instead of directly regressing $\mathbf{P}_0(t_j)$, we predict a residual track with respect to the reference frame:
\begin{equation}
\boldsymbol{\Delta}_j = \mathbf{P}_0(t_j) - \mathbf{P}_0(t_0).
\end{equation}
This residual formulation stabilizes training and improves accuracy (see Tab.~\ref{tab:ablation_design}), as $\boldsymbol{\Delta}_j = \mathbf{0}$ for static regions while non-zero values capture motion-induced displacement.

We decode $\hat{\mathbf{r}}_j$ using two separate VAE decoder heads:
\begin{equation}
\hat{\boldsymbol{\Delta}}_j = \mathcal{D}^{\text{track}}(\hat{\mathbf{r}}_j^{\Delta}), 
\qquad
\hat{\mathbf{o}}_j = \mathcal{D}^{\text{vis}}(\hat{\mathbf{r}}_j^{o}),
\end{equation}
where $\hat{\mathbf{r}}_j^{\Delta}$ and $\hat{\mathbf{r}}_j^{o}$ denote the channel-wise partitions. 
Here, $\hat{\boldsymbol{\Delta}}_j \in \mathbb{R}^{H \times W \times 3}$ is defined in the normalized pointmap space, and $\hat{\mathbf{o}}_j \in [0,1]^{H \times W}$ denotes visibility. 

Since the VAE decoder produces three-channel outputs, the visibility map is broadcast to three channels to match the output dimensionality~\cite{ke2024repurposing}. For pointmap normalization, we use the same factors (mean and maximum distance) as those of $\mathbf{P}_j(t_j)$ to ensure that the same physical point has the same 3D position after normalization.

Finally, the tracking pointmap is recovered as:
\begin{equation}
\hat{\mathbf{P}}_0(t_j) = \mathbf{P}_0(t_0) + \hat{\boldsymbol{\Delta}}_j.
\label{eq:residual}
\end{equation}

\paragrapht{Long-Video Inference.}
Our model is trained on clips of $1{+}F$ frames. To handle longer videos at inference time, we adopt a strided sliding window strategy with the first frame as a fixed anchor. 

Given a test video of $L$ frames, we compute the stride as $s = \lceil (L{-}1) / F \rceil$ and partition the frames $\{1, \ldots, L{-}1\}$ into $s$ non-overlapping groups. Each forward pass processes the anchor frame $\mathbf{I}_0$ together with $F$ frames sampled from one group, resulting in $s$ passes that cover the entire sequence. For each pass, we assign consecutive RoPE temporal indices $\{0, 1, \ldots, F\}$ as in training, regardless of the original frame indices. As in~\cite{harley2025alltracker, feng2025st4rtrack}, the model is trained with various temporal strides and naturally generalizes to non-consecutive frames. The predicted pointmaps are consistent across passes without post-processing, as all inputs $\mathbf{P}_j(t_j)$ share a common world coordinate frame. Fig.~\ref{fig:robustness} further evaluates the robustness of our method on long videos and large temporal strides.

\section{Experiment}
\subsection{Implementation Details}
\label{sec:impl_details}
\paragrapht{Architecture.} We fine-tune Wan 2.1-T2V~\cite{wan2025wan} using LoRA~\cite{hu2022lora}. Because the input and output token channel dimensions are doubled, we duplicate the DiT input projection weights~\cite{chefer2025videojam}. For the output projection, we retain the pre-trained weights for the first half of the channels and zero-initialize the remaining half. All VAE components are initialized from the pre-trained Wan VAE weights.

\paragrapht{Training.} All models are trained at a resolution of $480 \times 832$ on 12-frame clips using 8 H200 GPUs. Training proceeds in two stages. In Stage 1, we train the DiT with LoRA and input/output projection layers, with VAEs being frozen. We use AdamW~\cite{loshchilov2017decoupled} with a learning rate of $1\text{e-}4$ and a global batch size of 80 for 3 days. In Stage 2, we unfreeze all VAE encoders and decoders, $\mathcal{E}^{\text{rgb}}, \mathcal{E}^{\text{pm}}, \mathcal{D}^{\text{track}}, \mathcal{D}^{\text{vis}}$, and continue end-to-end training with learning rates of $3\text{e-}5$ for the DiT and $1\text{e-}5$ for the VAE, using a global batch size of 64 for an additional 2 days.

\paragrapht{Training Objective.} We minimize an MSE loss~\cite{ngo2025deltav2} on the predicted residual $\hat{\boldsymbol{\Delta}}_j$ in normalized pointmap space, combined with a BCE loss~\cite{karaev2025cotracker3} on visibility $\hat{\textbf{o}}_j$, weighted by 0.1.

\paragrapht{Dataset.} 
Following~\cite{feng2025st4rtrack}, we train our model on Kubric~\cite{greff2022kubric}, PointOdyssey~\cite{zheng2023pointodyssey}, and Dynamic Replica~\cite{karaev2023dynamicstereo}, which provide ground-truth 3D trajectories from mesh vertices. We also include TartanAir~\cite{wang2020tartanair}, a static-scene dataset with large camera motion, to improve robustness to ego-motion. More details are provided in Appendix~\ref{sec:appendix_dataset}.
 
\begin{table*}[t]
\centering
\small
\setlength{\tabcolsep}{3pt}
\renewcommand{\arraystretch}{1.05}
\caption{\textbf{3D tracking comparison.} We report AJ, $\text{APD}_\text{3D}$, and OA after Sim(3) alignment. The \colorbox{cvprblue!40}{best} and \colorbox{cvprblue!15}{second-best results} are highlighted in dark and light blue, respectively.}
\vspace{3pt}
\label{tab:main_comparison}
\resizebox{\textwidth}{!}{%
\begin{tabular}{l|ccc|ccc|ccc|ccc|ccc|ccc}
\toprule
 & \multicolumn{3}{c|}{ADT~\cite{pan2023aria}} & \multicolumn{3}{c|}{PStudio~\cite{joo2015panoptic}} & \multicolumn{3}{c|}{DR~\cite{karaev2023dynamicstereo}} & \multicolumn{3}{c|}{PO~\cite{zheng2023pointodyssey}} & \multicolumn{3}{c|}{Kubric~\cite{greff2022kubric}} & \multicolumn{3}{c}{Average} \\
\cmidrule(lr){2-4} \cmidrule(lr){5-7} \cmidrule(lr){8-10} \cmidrule(lr){11-13} \cmidrule(lr){14-16} \cmidrule(lr){17-19}
Method & AJ$\uparrow$ & $\text{APD}_\text{3D}$$\uparrow$ & OA$\uparrow$ & AJ$\uparrow$ & $\text{APD}_\text{3D}$$\uparrow$ & OA$\uparrow$ & AJ$\uparrow$ & $\text{APD}_\text{3D}$$\uparrow$ & OA$\uparrow$ & AJ$\uparrow$ & $\text{APD}_\text{3D}$$\uparrow$ & OA$\uparrow$ & AJ$\uparrow$ & $\text{APD}_\text{3D}$$\uparrow$ & OA$\uparrow$ & AJ$\uparrow$ & $\text{APD}_\text{3D}$$\uparrow$ & OA$\uparrow$ \\
\midrule

\multicolumn{19}{l}{\textit{(i) Iterative dense 3D trackers}} \\
DELTA~\cite{ngo2024delta} + ViPE~\cite{huang2025vipe}                                               &0.5088&	0.6949&	0.8142	&0.4987	&0.7810 &	0.6959	&0.4049	&0.5847	&0.7638	&0.4559&	0.6286&	0.8116	&0.2894&	0.3721&\cellcolor{cvprblue!45}0.9630&	0.4315 &	0.6123&	0.8097\\

DELTAv2~\cite{ngo2025deltav2} + ViPE~\cite{huang2025vipe}          
 &0.5135&	0.7066&	0.8038&	0.5353	&0.8026&	0.7284&	0.4167&	0.5888&	0.7825&	0.4459&	0.6246&	0.8011&	0.2860 &	0.3692&	0.9564&	0.4395 &	0.6184 &	0.8144\\
 DELTAv2~\cite{ngo2025deltav2} + DA3~\cite{lin2025depth}          
 &0.6150  &\cellcolor{cvprblue!15}0.8219&0.8125 &	0.5571 &\cellcolor{cvprblue!15}0.8496&	0.7087 &	0.4494 &	0.6217 &	0.7817	 &0.5304	 &0.7251	 &0.8020  &\cellcolor{cvprblue!15}0.3354	 &\cellcolor{cvprblue!15}0.4106 &	0.9592 &0.4975&\cellcolor{cvprblue!15}0.6858&	0.8128\\
\midrule
\multicolumn{19}{l}{\textit{(ii) Feed-forward dense 3D trackers based on 3D reconstruction models}} \\
St4RTrack~\cite{feng2025st4rtrack}                                                 
&0.5929	&0.7683	&0.8323&	0.5723	&0.7552	&0.8099&	0.3534	&0.5710 &	0.6836&	0.3968	&0.6579&	0.6860&	0.1193	&0.1896&	0.7703&	0.4069 &	0.5884	&0.7564 \\
Any4D~\cite{karhade2025any4d} 
& 0.4646&	0.6134	&0.8358	&0.4222	&0.5707	&0.8128	&0.4414&	0.6955&	0.6801&	0.4387	&0.6830&	0.7353&	0.3887&	0.4967	&0.8826&	0.4311 &	0.6119 &	0.7893 \\
TraceAnything~\cite{liu2025trace}                                              &0.5929&	0.7634	&0.8412&	0.5225&	0.6928&	0.8130&	0.2072	&0.3549&	0.7332&	0.2041	&0.3649	&0.6930&	0.2418	&0.3252&	0.8198&	0.3537&	0.5002 &	0.7800\\
\midrule
\multicolumn{19}{l}{\textit{(iii) Feed-forward dense 3D trackers based on video generative models}} \\
MotionCrafter~\cite{zhu2026motioncrafter}    
& 0.4460 &	0.6037&	0.8036	&0.5044	&0.6659	&0.8142&	0.4926&	0.6172	&0.9172	&0.4197&	0.6412	&0.7301&	0.2176&	0.3011	&0.8730 &	0.4161 &	0.5658 &	0.8276\\

\textbf{\ours} + ViPE~\cite{huang2025vipe}                                   
&\cellcolor{cvprblue!15}0.6683&	0.7688&\cellcolor{cvprblue!15}0.9405&\cellcolor{cvprblue!15}0.6803&0.8163&\cellcolor{cvprblue!45}0.8937&\cellcolor{cvprblue!15}0.5842&\cellcolor{cvprblue!15}0.7034&\cellcolor{cvprblue!45}0.9408&\cellcolor{cvprblue!15}0.5836	&\cellcolor{cvprblue!15}0.7262&\cellcolor{cvprblue!45}0.8943&	0.3032&	0.3940&	\cellcolor{cvprblue!15}0.9597&\cellcolor{cvprblue!15}0.5639&	0.6817 &\cellcolor{cvprblue!45}0.9258\\
\textbf{\ours} + DA3~\cite{lin2025depth}                                   
&\cellcolor{cvprblue!45}0.8626&\cellcolor{cvprblue!45}0.9510&\cellcolor{cvprblue!45}0.9445&\cellcolor{cvprblue!45}0.7287&\cellcolor{cvprblue!45}0.8713&\cellcolor{cvprblue!15}0.8891&\cellcolor{cvprblue!45}0.6518&\cellcolor{cvprblue!45}0.7706&\cellcolor{cvprblue!15}0.9388&\cellcolor{cvprblue!45}0.7288&	\cellcolor{cvprblue!45}0.8678&\cellcolor{cvprblue!15}0.8938&\cellcolor{cvprblue!45}0.4208&\cellcolor{cvprblue!45}0.5047&0.9587&\cellcolor{cvprblue!45}0.6785&\cellcolor{cvprblue!45}0.7931&\cellcolor{cvprblue!15}0.9250\\

\bottomrule
\end{tabular}%
}
\end{table*}

\begin{figure*}[t]
  \centering
  \includegraphics[width=\textwidth]{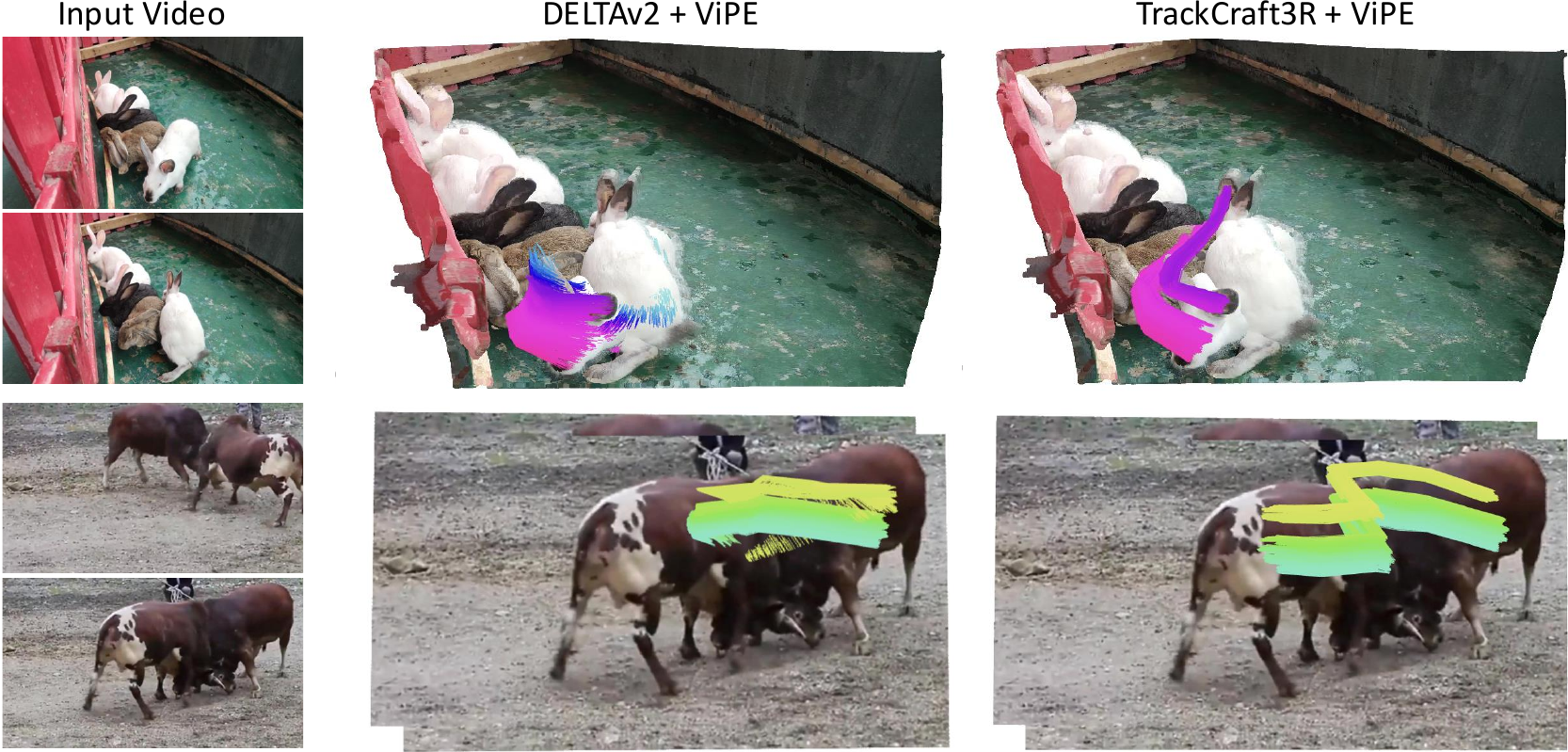}
\vspace{-15pt}
\caption{\textbf{Qualitative comparison on ITTO~\cite{demler2025tracker} videos.} \ours\ accurately estimates dense 3D trajectories on real-world videos under large object dynamics and occlusion.}
  \label{fig:qualitative}\vspace{-10pt}
\end{figure*}

\subsection{Evaluation Settings}
\label{sec:eval_settings}

\paragrapht{Evaluation Datasets.}
We evaluate our method on both 3D sparse and dense tracking benchmarks, with all metrics computed in a world coordinate frame. For \emph{3D sparse tracking}, following~\cite{feng2025st4rtrack}, we use two real-world datasets from TAPVid-3D~\cite{koppula2024tapvid}, Aerial Digital Twin (ADT)~\cite{pan2023aria} and Panoptic Studio (PStudio)~\cite{joo2015panoptic}, along with two synthetic test dataset, Point Odyssey (PO)~\cite{zheng2023pointodyssey} and Dynamic Replica (DR)~\cite{karaev2023dynamicstereo}. Each dataset provides sparse ground-truth 3D trajectories, and we evaluate the first 84 frames. For \emph{3D dense tracking}, following~\cite{ngo2024delta,ngo2025deltav2, karhade2025any4d}, we use the held-out Kubric~\cite{greff2022kubric} test split consisting of 50 sequences. This dataset provides dense ground-truth 3D trajectories defined for every pixel in the reference frame, and we evaluate the first 24 frames following~\cite{ngo2025deltav2}.

\paragrapht{Evaluation Metrics.}
Following TAPVid-3D~\cite{koppula2024tapvid}, we report three metrics: (i) \emph{average percentage of points within $\delta_{3\mathrm{D}}$} ($\text{APD}_\text{3D}$), defined as the percentage of points whose 3D end-point error is below a threshold $\delta_{3\mathrm{D}} \in \{0.1, 0.3, 0.5, 1\}\mathrm{m}$~\cite{feng2025st4rtrack}, averaged over thresholds; (ii) \emph{occlusion accuracy} (OA), which measures the accuracy of occlusion prediction as a binary classification task; and (iii) \emph{average Jaccard} (AJ), which jointly measures 3D point accuracy and occlusion prediction. Predicted trajectories are aligned to the ground truth using Sim(3) alignment~\cite{xiao2025spatialtrackerv2, feng2025st4rtrack}.

\paragrapht{Baselines.}
We compare our method against recent dense 3D trackers, grouped into three categories. \emph{(i) Iterative Dense 3D Trackers}: DELTA~\cite{ngo2024delta} and DELTAv2~\cite{ngo2025deltav2}, which condition on external depth and use camera poses to transform tracks into world coordinates. \emph{(ii) Feed-forward Dense 3D Trackers Based on 3D Reconstruction Models}: St4RTrack~\cite{feng2025st4rtrack}, Any4D~\cite{karhade2025any4d}, and TraceAnything~\cite{liu2025trace} are built upon pre-trained 3D reconstruction backbones~\cite{leroy2024grounding, keetha2025mapanything, yang2025fast3r}, which are pre-trained to predict camera poses, depth, or pointmaps. \emph{(iii) Feed-forward Dense 3D Trackers Based on Video Generative Models}: MotionCrafter~\cite{zhu2026motioncrafter}, based on a video diffusion U-Net~\cite{blattmann2023stable}, where tracks are obtained by chaining scene flow across adjacent frames. Since \emph{(ii)} does not output visibility, we project the predicted track pointmaps into each frame and consider a point visible if the projected pixel lies within the image bounds and its projected depth is within a 10\% tolerance of the per-frame depth. We use ViPE~\cite{huang2025vipe} and DA3~\cite{lin2025depth} to provide input geometry for DELTA, DELTAv2, and \ours.

\begin{figure*}[!t]
  \centering
  \setlength{\tabcolsep}{1pt}
  \begin{tabular}{@{}c@{\hspace{0pt}}c@{\hspace{0pt}}c@{\hspace{0pt}}c@{}}
    \includegraphics[width=0.25\textwidth]{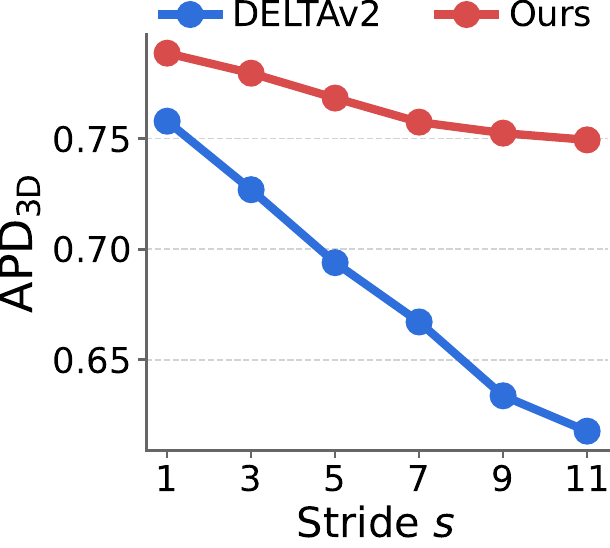} &
    \includegraphics[width=0.24\textwidth]{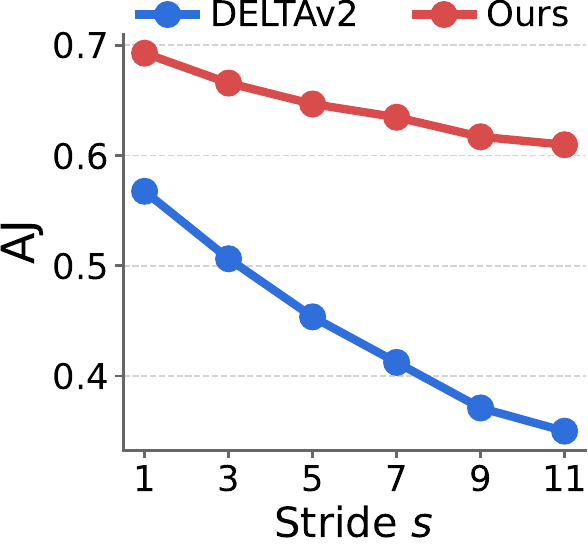} &
    \includegraphics[width=0.25\textwidth]{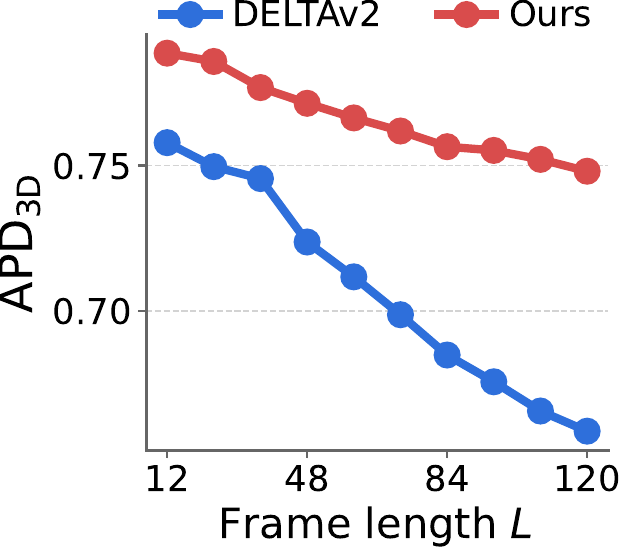} &
    \includegraphics[width=0.24\textwidth]{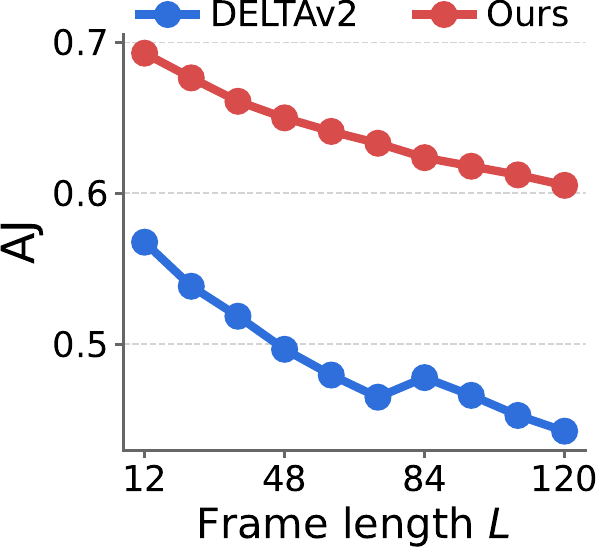} \\
    {\small (a) $\text{APD}_\text{3D}$, varying $s$} &
    {\small (b) AJ, varying $s$} &
    {\small (c) $\text{APD}_\text{3D}$, varying $L$} &
    {\small (d) AJ, varying $L$} \\
  \end{tabular}
  \vspace{-5pt}
  \caption{\textbf{Robustness to large inter-frame motion (a, b) and long videos (c, d).} \textbf{\textcolor{Red}{\ours}}'s performance drops much more slowly than \textbf{\textcolor{Blue}{DELTAv2}} as stride $s$ or frame length $L$ grows.}
  \label{fig:robustness}\vspace{-10pt}
\end{figure*}

\paragrapht{Quantitative Comparison.} In Tab.~\ref{tab:main_comparison}, \ours\ achieves state-of-the-art performance across all benchmarks, with the best average AJ, $\text{APD}_\text{3D}$, and OA. \ours\ + ViPE surpasses the strongest iterative dense 3D tracker, DELTAv2~\cite{ngo2025deltav2} + ViPE, as well as all feed-forward baselines. \ours\ + ViPE is even competitive with DELTAv2~\cite{ngo2025deltav2} + DA3. With the stronger 3D foundation model DA3, \ours\ + DA3 further surpasses DELTAv2 + DA3 and outperforms all other feed-forward baselines by a large margin. More quantitative comparisons with a dense 2D tracker~\cite{harley2025alltracker}, sparse 3D trackers~\cite{xiao2025spatialtrackerv2, zhang2025tapip3d}, and the concurrent work V-DPM~\cite{sucar2026v} are provided in Appendix Secs.~\ref{appendix:alltracker},~\ref{sec:appendix_sparse}, and~\ref{sec:appendix_vdpm}.

\paragrapht{Qualitative Comparison.}
Fig.~\ref{fig:qualitative} compares 3D trajectories predicted by \ours\ and 
DELTAv2~\cite{ngo2025deltav2} on real-world ITTO~\cite{demler2025tracker} videos. 
\ours\ produces accurate dense trajectories under large object dynamics and 
occlusion, where DELTAv2 often fails. Additional results on ITTO~\cite{demler2025tracker} and DAVIS~\cite{pont20172017} 
are provided in Appendix Sec.~\ref{sec:appendix_qualitative} and the supplementary video.

\paragrapht{Robustness to Large Motion and Long Videos.}
We further evaluate the robustness of our method with respect to \textit{motion} and \textit{video length}. For \emph{large motion}, we fix the clip length to 12 frames and increase the temporal stride $s$ from 1 to 12 (in steps of 1), enlarging per-frame displacement. For \emph{long videos}, we fix the stride to $s{=}1$ and increase the sequence length $L$ from 12 to 120 (in steps of 12). The resulting $\text{APD}_\text{3D}$ and AJ curves are shown in Fig.~\ref{fig:robustness}, averaged over sparse tracking benchmarks. \ours\ consistently widens the gap with DELTAv2~\cite{ngo2025deltav2} in both settings, indicating that the learned motion prior enables robust tracking under large displacements and generalizes to long-horizon videos beyond the training sequence length (12 frames).

\subsection{Ablation Study}
\label{sec:ablation}

\begin{wraptable}[3]{r}{0.37\textwidth}
\centering
\vspace{-40pt}
\caption{\textbf{Ablation on spatio-temporal priors.}}
\vspace{3pt}
\label{tab:ablation_pretrained}
\resizebox{\linewidth}{!}{
\begin{tabular}{l|ccc}
\toprule
Initialization & AJ$\uparrow$ & $\text{APD}_\text{3D}$$\uparrow$ & OA$\uparrow$ \\
\midrule
Random         &  0.4698	& 0.6312 &	0.8271 \\
Pre-trained (Ours)    & \textbf{0.5639}	& \textbf{0.6817} &	\textbf{0.9258} \\
\bottomrule

\end{tabular}
}
\end{wraptable}

All ablation studies report the average AJ, $\text{APD}_\text{3D}$, and OA across all benchmarks, and use ViPE~\cite{huang2025vipe} for input geometry, unless otherwise specified. 

\paragrapht{Spatio-Temporal Prior in Video DiT.} 
We compare our model against an identical architecture trained from scratch, both to convergence on the same data. Tab.~\ref{tab:ablation_pretrained} shows that random initialization substantially degrades all metrics, confirming that the pre-trained spatio-temporal prior is critical.

\begin{wraptable}[7]{r}{0.45\textwidth}
\centering
\vspace{-18pt}
\caption{\textbf{Ablation on model components.}}
\vspace{3pt}
\label{tab:ablation_design}
\resizebox{\linewidth}{!}{
\begin{tabular}{l|ccc}
\toprule
Configuration & AJ$\uparrow$ & $\text{APD}_\text{3D}$$\uparrow$ & OA$\uparrow$ \\
\midrule
(a) w/o First-frame anchoring   & 0.5135 &	0.6535  &	0.8778\\
(b) w/o Temporal RoPE alignment & 0.4450	& 0.6317 & 	0.8031 \\
(c) w/o Residual displacement   & 0.5007&	0.6172 &	0.9159\\
(d) w/ VAE temporal compression & 0.4487&	0.6325	&0.8148\\
\midrule
Full model (Ours)             &\textbf{0.5609}	&\textbf{0.6790} &	\textbf{0.9225} \\
\bottomrule
\end{tabular}
}
\end{wraptable}

\paragrapht{Model Design.} Tab.~\ref{tab:ablation_design} ablates four core design components by removing each from the full model. For this ablation, we train each model with the VAE frozen. \emph{(a) w/o First-frame anchoring}: setting $\mathbf{r}_j = \mathbf{g}_j$ instead of $\mathbf{r}_j = \mathbf{g}_0$, removing reference-frame anchoring. \emph{(b) w/o Temporal RoPE alignment}: assigning a constant temporal index $t_0$ to all $\mathbf{r}_j$. \emph{(c) w/o Residual displacement}: directly regressing the 3D track pointmap $\mathbf{P}_0(t_j)$ instead of predicting residual displacements ${\boldsymbol{\Delta}}_j$. \emph{(d) w/ VAE temporal compression}: using the original 3D VAE temporal downsampling instead of processing frames independently. All four components contribute. (a) causes consistent drops across all metrics and (b) causes the largest AJ drop, indicating that (a) and (b) jointly contribute to reference-anchored correspondence at the correct target timestamp (Fig.~\ref{fig:attention}). (c) specifically drops $\text{APD}_\text{3D}$, since residual prediction stabilizes pointmap prediction. (d) drops all metrics consistently, as VAE temporal compression affects both the pointmap and visibility decoders.

\begin{wraptable}[6]{r}{0.40\textwidth}
\centering
\vspace{-18pt}
\caption{\textbf{Ablation on input geometry.}}
\vspace{3pt}
\label{tab:ablation_geometry}
\resizebox{\linewidth}{!}{
\begin{tabular}{l|ccc}
\toprule
Configuration & AJ$\uparrow$ & $\text{APD}_\text{3D}$$\uparrow$ & OA$\uparrow$ \\
\midrule
DELTAv2~\cite{ngo2025deltav2} + DA3~\cite{lin2025depth} &  0.4384	&0.5858&	0.8476\\
\textbf{\ours} + DA3~\cite{lin2025depth} & \textbf{0.6005} & \textbf{0.7144}& 	\textbf{0.9304}\\
\midrule
DELTAv2~\cite{ngo2025deltav2} + GT   & 0.5590 & 0.7169 & 0.8380 \\
\textbf{\ours} + GT   & \textbf{0.7649} & \textbf{0.8635} & \textbf{0.9353} \\
\bottomrule
\end{tabular}
}
\end{wraptable}
\paragrapht{Input Geometry Quality.}
In Tab.~\ref{tab:ablation_geometry}, we study the impact of input geometry quality by using depth and camera poses from DA3~\cite{lin2025depth} and ground truth (GT). Metrics are averaged over the synthetic datasets~\cite{zheng2023pointodyssey, karaev2023dynamicstereo, greff2022kubric}, for which GT is available. Without any retraining, replacing DA3 with GT consistently improves all metrics, providing an upper bound for our method and suggesting that future advances in 3D geometry estimation can directly translate to better tracking performance. Note that performing tracking with input geometry from off-the-shelf estimators is becoming a common practice~\cite{xiao2025spatialtrackerv2, zhang2025tapip3d, ngo2024delta, ngo2025deltav2}. In Appendix Sec.~\ref{sec:appendix_sparse}, we further compare \ours\ with recent sparse 3D trackers~\cite{xiao2025spatialtrackerv2, zhang2025tapip3d} under the same input geometry.

\begin{wraptable}[6]{r}{0.40\textwidth}
\centering
\vspace{-20pt}
\caption{\textbf{Ablation on LoRA rank and VAE finetuning.}}
\vspace{3pt}
\label{tab:ablation_scaling_lora}
\resizebox{\linewidth}{!}{
\begin{tabular}{cc|ccc}
\toprule
LoRA rank & VAE finetuning & AJ$\uparrow$ & $\text{APD}_\text{3D}$$\uparrow$ & OA$\uparrow$ \\
\midrule
64   & \ding{55}   & 0.5025 &	0.6430 &	0.8779 \\
256  & \ding{55}   & 0.5399 &	0.6623 &	0.9112 \\
1024 & \ding{55}   &0.5609	&0.6790 &	0.9225\\
1024 & \ding{51}   & \textbf{0.5639}	& \textbf{0.6817} &	\textbf{0.9258} \\
\bottomrule
\end{tabular}
}
\end{wraptable}

\paragrapht{LoRA Rank and VAE Finetuning.} In Tab.~\ref{tab:ablation_scaling_lora}, increasing the LoRA rank from 64 to 1024 consistently improves performance, indicating that the DiT benefits from more expressive low-rank updates. Unfreezing the VAEs in Stage 2 yields further gains, confirming the benefit of adapting them to the pointmap and visibility domains.

\begin{wraptable}[9]{r}{0.40\textwidth}
\centering
\vspace{-15pt}
\caption{\textbf{Inference efficiency} with different frame lengths.}
\vspace{3pt}
\label{tab:efficiency}
\resizebox{\linewidth}{!}{
\begin{tabular}{c|lcc}
\toprule
Frames & Method & Time (s)$\downarrow$ & Memory (GB)$\downarrow$ \\
\midrule
12 & DELTA~\cite{ngo2024delta}     &   14.64  & 29.97 \\
12 & DELTAv2~\cite{ngo2025deltav2} &  5.00  & 35.46 \\
12 & \textbf{\ours}                        & \textbf{  3.91   }  & \textbf{7.63} \\
\midrule
23 & DELTA~\cite{ngo2024delta}     & 28.92 & 30.78 \\
23 & DELTAv2~\cite{ngo2025deltav2} & 9.70        & 35.90 \\
23 & \textbf{\ours}                   & \textbf{7.84}  & \textbf{7.63} \\
\bottomrule
\end{tabular}
}
\end{wraptable}
\paragrapht{Inference Efficiency.}
Tab.~\ref{tab:efficiency} compares inference time and peak GPU memory of \ours, DELTA~\cite{ngo2024delta}, and DELTAv2~\cite{ngo2025deltav2} at $448{\times}448$ resolution for 12- and 23-frame clips on a single NVIDIA A6000 GPU. For 12 frames, \ours\ is \textbf{1.3$\times$ faster} and uses \textbf{4.6$\times$ less peak memory} than DELTAv2. Specifically, DELTA and DELTAv2 (1) perform iterative refinement (six steps) and (2) construct 4D correlation features between dense queries and multi-scale image features. In contrast, \ours\ (1) predicts trajectories in a \emph{single forward pass} and (2) replaces explicit 4D correlation features with full 3D attention in a \emph{$1/16$ spatially compressed latent space}, which is effectively upsampled to pixel space with a VAE decoder. For longer sequences (\eg, 23 frames), the same trend holds: all methods scale roughly linearly in runtime, while peak memory remains similar.

\section{Conclusion}
\vspace{-5pt}
We presented \ours, the first method to repurpose a video diffusion 
transformer as a single-pass dense 3D tracker. By introducing a 
dual-latent representation that couples per-frame geometry latents with 
first-frame-anchored track latents, together with a temporal RoPE alignment 
that specifies the target timestamp of each track latent, \ours\ converts 
the per-frame generative paradigm of video DiTs into a reference-anchored 
dense tracking paradigm with LoRA fine-tuning. \ours\ 
achieves state-of-the-art performance on standard 3D sparse and dense 
tracking benchmarks while running $1.3{\times}$ faster and using 
$4.6{\times}$ less peak memory than the strongest iterative 3D tracker, 
and further demonstrates robustness to large motions and long videos.

\clearpage

{
\small
\bibliographystyle{plainnat}
\bibliography{references}
}

%%%%%%%%%%%%%%%%%%%%%%%%%%%%%%%%%%%%%%%%%%%%%%%%%%%%%%%%%%%%
\clearpage

\appendix
This appendix complements the main paper with the following:
\begin{itemize}
    \item Sec.~\ref{sec:appendix_dataset}: Additional details on the training dataset.
    \item Sec.~\ref{appendix:alltracker}: Comparison with a dense 2D tracker~\cite{harley2025alltracker}.
    \item Sec.~\ref{sec:appendix_sparse}: Comparison with sparse 3D trackers~\cite{xiao2025spatialtrackerv2, zhang2025tapip3d}.
    \item Sec.~\ref{sec:appendix_vdpm}: Comparison with V-DPM~\cite{sucar2026v}.
    \item Sec.~\ref{sec:appendix_attention}: Attention visualizations.
    \item Sec.~\ref{sec:appendix_qualitative}: Additional qualitative results.
    \item Sec.~\ref{sec:appendix_limitation}: Limitations and future work.
\end{itemize}

\begin{wraptable}{r}{0.42\textwidth}
\centering
\vspace{-15pt}
\caption{\textbf{Training data.} Number of videos and temporal strides.}
\vspace{3pt}
\label{tab:training_data}
\resizebox{\linewidth}{!}{
\begin{tabular}{l|cc}
\toprule
Dataset & \#Videos & Strides \\
\midrule
Kubric~\cite{greff2022kubric}                 & 6{,}042 & [3, 4, 5, 6, 7] \\
DynamicReplica~\cite{karaev2023dynamicstereo} & 483     & [5, 6, 7, 8, 9] \\
PointOdyssey~\cite{zheng2023pointodyssey}     & 45      & [2, 3, 4, 5, 6] \\
TartanAir~\cite{wang2020tartanair}            & 163     & [1, 2, 3] \\
\bottomrule
\end{tabular}
}
\vspace{-10pt}
\end{wraptable}
\section{Training Datasets}
\label{sec:appendix_dataset}
As summarized in Tab.~\ref{tab:training_data}, we train \ours\ on four synthetic datasets: Kubric~\cite{greff2022kubric}, DynamicReplica~\cite{karaev2023dynamicstereo}, PointOdyssey~\cite{zheng2023pointodyssey}, and TartanAir~\cite{wang2020tartanair}. Kubric, DynamicReplica, and PointOdyssey provide RGB, depth, camera parameters, and 3D trajectories. For Kubric~\cite{greff2022kubric}, following~\cite{feng2025st4rtrack}, we render 6K sequences ($480{\times}832$, 81 frames) and extract dense trajectories from the first frame. DynamicReplica and PointOdyssey provide sparse 3D trajectories from mesh vertices. TartanAir contains static scenes with large camera motion and provides RGB, depth, and camera poses. During training, we randomly sample a temporal stride from the strides listed in Tab.~\ref{tab:training_data} for each dataset to cover diverse motion patterns. 

\section{Comparison with Lifted Dense 2D Tracker}
\vspace{-15pt}
\begin{table*}[h]
\label{appendix:alltracker}
\centering
\small
\setlength{\tabcolsep}{3pt}
\renewcommand{\arraystretch}{1.05}
\caption{\textbf{3D tracking comparison with lifted AllTracker~\cite{harley2025alltracker}.} We report AJ, $\text{APD}_\text{3D}$, and OA after Sim(3) alignment. The estimated dense 2D tracks from AllTracker are lifted to 3D using ViPE~\cite{huang2025vipe} depth and camera poses. The \colorbox{cvprblue!40}{best} and \colorbox{cvprblue!15}{second-best results} are highlighted in dark and light blue, respectively.}
\vspace{3pt}
\label{tab:alltracker_comparison}
\resizebox{\textwidth}{!}{%
\begin{tabular}{l|ccc|ccc|ccc|ccc|ccc|ccc}
\toprule
 & \multicolumn{3}{c|}{ADT~\cite{pan2023aria}} 
 & \multicolumn{3}{c|}{PStudio~\cite{joo2015panoptic}} 
 & \multicolumn{3}{c|}{DR~\cite{karaev2023dynamicstereo}} 
 & \multicolumn{3}{c|}{PO~\cite{zheng2023pointodyssey}} 
 & \multicolumn{3}{c|}{Kubric~\cite{greff2022kubric}} 
 & \multicolumn{3}{c}{Average} \\
\cmidrule(lr){2-4} \cmidrule(lr){5-7} \cmidrule(lr){8-10} \cmidrule(lr){11-13} \cmidrule(lr){14-16} \cmidrule(lr){17-19}
Method 
& AJ$\uparrow$ & $\text{APD}_\text{3D}$$\uparrow$ & OA$\uparrow$
& AJ$\uparrow$ & $\text{APD}_\text{3D}$$\uparrow$ & OA$\uparrow$
& AJ$\uparrow$ & $\text{APD}_\text{3D}$$\uparrow$ & OA$\uparrow$
& AJ$\uparrow$ & $\text{APD}_\text{3D}$$\uparrow$ & OA$\uparrow$
& AJ$\uparrow$ & $\text{APD}_\text{3D}$$\uparrow$ & OA$\uparrow$
& AJ$\uparrow$ & $\text{APD}_\text{3D}$$\uparrow$ & OA$\uparrow$ \\
\midrule

AllTracker~\cite{harley2025alltracker} + ViPE~\cite{huang2025vipe} 
&\cellcolor{cvprblue!15}0.6177&\cellcolor{cvprblue!15}0.7234&\cellcolor{cvprblue!15}0.9421&\cellcolor{cvprblue!15}0.6463&\cellcolor{cvprblue!45}0.8214&\cellcolor{cvprblue!15}0.8301&\cellcolor{cvprblue!15}0.4719&\cellcolor{cvprblue!15}0.5979&\cellcolor{cvprblue!15}0.9123&\cellcolor{cvprblue!15}0.5410&\cellcolor{cvprblue!15}0.7098&\cellcolor{cvprblue!15}0.8601&\cellcolor{cvprblue!15}0.2903&\cellcolor{cvprblue!15}0.3879&\cellcolor{cvprblue!15}0.9318&\cellcolor{cvprblue!15}0.5134&\cellcolor{cvprblue!15}0.6481 &\cellcolor{cvprblue!15}0.8953\\

\textbf{\ours} + ViPE~\cite{huang2025vipe}                                   
&\cellcolor{cvprblue!45}0.6683&	\cellcolor{cvprblue!45}0.7688&\cellcolor{cvprblue!45}0.9405&\cellcolor{cvprblue!45}0.6803&\cellcolor{cvprblue!15}0.8163&\cellcolor{cvprblue!45}0.8937&\cellcolor{cvprblue!45}0.5842&\cellcolor{cvprblue!45}0.7034&\cellcolor{cvprblue!45}0.9408&\cellcolor{cvprblue!45}0.5836	&\cellcolor{cvprblue!45}0.7262&\cellcolor{cvprblue!45}0.8943&\cellcolor{cvprblue!45}0.3032&\cellcolor{cvprblue!45}0.3940&	\cellcolor{cvprblue!45}0.9597&\cellcolor{cvprblue!45}0.5639&\cellcolor{cvprblue!45}0.6817 &\cellcolor{cvprblue!45}0.9258\\

\bottomrule
\end{tabular}%
}
% \vspace{-20pt}
\end{table*}
In Tab.~\ref{tab:alltracker_comparison}, we further compare \ours\ with AllTracker~\cite{harley2025alltracker}, a recent dense 2D tracker, on sparse and dense 3D tracking benchmarks. We use depth and camera pose from ViPE~\cite{huang2025vipe} to unproject the estimated 2D tracks into 3D world coordinates. \ours\ consistently outperforms AllTracker, achieving higher overall AJ, $\text{APD}_\text{3D}$, and OA across all benchmarks.

\section{Comparison with Sparse 3D Trackers}
\vspace{-15pt}
\begin{table*}[h]
\centering
\small
\setlength{\tabcolsep}{3pt}
\renewcommand{\arraystretch}{1.05}
\caption{\textbf{3D tracking comparison with SpatialTrackerV2~\cite{xiao2025spatialtrackerv2} and TAPIP3D~\cite{zhang2025tapip3d}.} We report AJ, $\text{APD}_\text{3D}$, and OA after Sim(3) alignment. The \colorbox{cvprblue!40}{best} and \colorbox{cvprblue!15}{second-best results} are highlighted in dark and light blue, respectively.}
\vspace{3pt}
\label{tab:sparse_comparison}
\resizebox{\textwidth}{!}{
\begin{tabular}{l|ccc|ccc|ccc|ccc|ccc}
\toprule
 & \multicolumn{3}{c|}{ADT~\cite{pan2023aria}} 
 & \multicolumn{3}{c|}{PStudio~\cite{joo2015panoptic}} 
 & \multicolumn{3}{c|}{DR~\cite{karaev2023dynamicstereo}} 
 & \multicolumn{3}{c|}{PO~\cite{zheng2023pointodyssey}} 
 & \multicolumn{3}{c}{Average} \\
\cmidrule(lr){2-4} \cmidrule(lr){5-7} \cmidrule(lr){8-10} \cmidrule(lr){11-13} \cmidrule(lr){14-16}
Method 
& AJ$\uparrow$ & $\text{APD}_\text{3D}$$\uparrow$ & OA$\uparrow$
& AJ$\uparrow$ & $\text{APD}_\text{3D}$$\uparrow$ & OA$\uparrow$
& AJ$\uparrow$ & $\text{APD}_\text{3D}$$\uparrow$ & OA$\uparrow$
& AJ$\uparrow$ & $\text{APD}_\text{3D}$$\uparrow$ & OA$\uparrow$
& AJ$\uparrow$ & $\text{APD}_\text{3D}$$\uparrow$ & OA$\uparrow$ \\
\midrule
SpatialTrackerV2~\cite{xiao2025spatialtrackerv2} + ViPE~\cite{huang2025vipe}
&0.6533&\cellcolor{cvprblue!45}0.7818&	0.9193&	0.6152&	0.7961	&0.8142&\cellcolor{cvprblue!15}0.4499&0.5762&\cellcolor{cvprblue!15}0.9172&	\cellcolor{cvprblue!15}0.5023	&\cellcolor{cvprblue!15}0.7060&	0.8042&	0.5552&	0.7150&\cellcolor{cvprblue!15}0.8637\\
TAPIP3D~\cite{zhang2025tapip3d} + ViPE~\cite{huang2025vipe}
&\cellcolor{cvprblue!15}0.6616	&0.7602&\cellcolor{cvprblue!45}0.9419&\cellcolor{cvprblue!15}0.6426&	\cellcolor{cvprblue!45}0.8224	&\cellcolor{cvprblue!15}0.8248&0.4131&\cellcolor{cvprblue!15}0.5925&0.7586 &	0.5487	&0.7055	&\cellcolor{cvprblue!15}0.8698&\cellcolor{cvprblue!15}0.5665&\cellcolor{cvprblue!15}0.7202 &0.8488 \\

\textbf{\ours} + ViPE~\cite{huang2025vipe}                                   
&\cellcolor{cvprblue!45}0.6683&\cellcolor{cvprblue!15}0.7688&\cellcolor{cvprblue!15}0.9405	&\cellcolor{cvprblue!45}0.6803&\cellcolor{cvprblue!15}0.8163&\cellcolor{cvprblue!45}0.8937&\cellcolor{cvprblue!45}0.5842&\cellcolor{cvprblue!45}0.7034&\cellcolor{cvprblue!45}0.9408&\cellcolor{cvprblue!45}0.5836&\cellcolor{cvprblue!45}0.7262&\cellcolor{cvprblue!45}0.8943&\cellcolor{cvprblue!45}0.6291&	\cellcolor{cvprblue!45}0.7537 &\cellcolor{cvprblue!45}0.9173\\
\bottomrule
\end{tabular}
}
\end{table*}
\label{sec:appendix_sparse}
In Tab.~\ref{tab:sparse_comparison}, we compare \ours\ with the recent sparse 3D trackers SpatialTrackerV2~\cite{xiao2025spatialtrackerv2} and TAPIP3D~\cite{zhang2025tapip3d} on the sparse 3D tracking benchmarks. Note that SpatialTrackerV2 and TAPIP3D also take camera poses and depth from off-the-shelf models as input. For fair comparison, we use ViPE~\cite{huang2025vipe} for all methods. \ours\ outperforms both SpatialTrackerV2 and TAPIP3D, achieving the best average AJ, $\text{APD}_\text{3D}$, and OA.

\section{Comparison with V-DPM}
\label{sec:appendix_vdpm}

\begin{table*}[t]
\centering
\small
\setlength{\tabcolsep}{3pt}
\renewcommand{\arraystretch}{1.05}
\caption{\textbf{3D tracking comparison with V-DPM~\cite{sucar2026v}.} We report AJ, $\text{APD}_\text{3D}$, and OA after Sim(3) alignment. The \colorbox{cvprblue!40}{best} and \colorbox{cvprblue!15}{second-best results} are highlighted in dark and light blue, respectively.}
\vspace{3pt}
\label{tab:vdpm_comparison}
\resizebox{\textwidth}{!}{%
\begin{tabular}{l|ccc|ccc|ccc|ccc|ccc}
\toprule
 & \multicolumn{3}{c|}{ADT~\cite{pan2023aria}} 
 & \multicolumn{3}{c|}{PStudio~\cite{joo2015panoptic}} 
 & \multicolumn{3}{c|}{DR~\cite{karaev2023dynamicstereo}} 
 & \multicolumn{3}{c|}{PO~\cite{zheng2023pointodyssey}} 
 & \multicolumn{3}{c}{Average} \\
\cmidrule(lr){2-4} \cmidrule(lr){5-7} \cmidrule(lr){8-10} \cmidrule(lr){11-13} \cmidrule(lr){14-16}
Method 
& AJ$\uparrow$ & $\text{APD}_\text{3D}$$\uparrow$ & OA$\uparrow$
& AJ$\uparrow$ & $\text{APD}_\text{3D}$$\uparrow$ & OA$\uparrow$
& AJ$\uparrow$ & $\text{APD}_\text{3D}$$\uparrow$ & OA$\uparrow$
& AJ$\uparrow$ & $\text{APD}_\text{3D}$$\uparrow$ & OA$\uparrow$
& AJ$\uparrow$ & $\text{APD}_\text{3D}$$\uparrow$ & OA$\uparrow$ \\
\midrule
V-DPM~\cite{sucar2026v} 
&0.7745&\cellcolor{cvprblue!15}0.9126&	0.8668	&\cellcolor{cvprblue!45}0.8501&\cellcolor{cvprblue!45}0.9606	&0.9065	&0.6041	&\cellcolor{cvprblue!45}0.8319&0.7670&\cellcolor{cvprblue!15}0.7030&\cellcolor{cvprblue!45}0.9567	&0.7785&	0.7329 &\cellcolor{cvprblue!45}0.9155 &	0.8297\\
\midrule
\textbf{\ours} + DA3~\cite{lin2025depth}                                   
&\cellcolor{cvprblue!15}0.8040&	0.8692&\cellcolor{cvprblue!45}0.9701&\cellcolor{cvprblue!15}0.8190&\cellcolor{cvprblue!15}0.9036&\cellcolor{cvprblue!45}0.9454&\cellcolor{cvprblue!15}0.6575&	0.7614&\cellcolor{cvprblue!45}0.9742&	0.6835&	0.7902	&\cellcolor{cvprblue!45}0.9227	&\cellcolor{cvprblue!15}0.7410 &	0.8311&\cellcolor{cvprblue!45}0.9531\\
\textbf{\ours} + V-DPM~\cite{sucar2026v}                                   
&\cellcolor{cvprblue!45}0.8418&\cellcolor{cvprblue!45}0.9183&\cellcolor{cvprblue!15}0.9477&0.7518&	0.8607&\cellcolor{cvprblue!15}0.9227	&\cellcolor{cvprblue!45}0.7128	&\cellcolor{cvprblue!15}0.8150&\cellcolor{cvprblue!15}0.9626&\cellcolor{cvprblue!45}0.8146	&\cellcolor{cvprblue!15}0.9386	&\cellcolor{cvprblue!15}0.9008&\cellcolor{cvprblue!45}0.7803 &\cellcolor{cvprblue!15}0.8832 &\cellcolor{cvprblue!15}0.9335
\\
\bottomrule
\end{tabular}%
}
\end{table*}

While V-DPM~\cite{sucar2026v} is a concurrent work trained on different dataset scales, we provide additional comparisons for completeness. Tab.~\ref{tab:vdpm_comparison} reports AJ, $\text{APD}_\text{3D}$, and OA on sparse tracking benchmarks, evaluated on the first 24 frames. We also report \ours\ + V-DPM, which uses V-DPM's predicted frame-anchored reconstruction pointmaps as our input geometry. Both \ours\ + DA3~\cite{lin2025depth} and \ours\ + V-DPM outperform V-DPM in AJ and OA, while V-DPM achieves slightly higher $\text{APD}_\text{3D}$. Notably, \ours\ runs \textbf{6.6$\times$ faster} with \textbf{2.3$\times$ less memory} than V-DPM (Tab.~\ref{tab:vdpm_efficiency}). Below, we provide a detailed discussion of \ours\ \emph{vs.}\ V-DPM.

\paragrapht{Dataset Scale.}
V-DPM relies heavily on 3D/4D supervision throughout its training. Its backbone (VGGT~\cite{wang2025vggt}) is pre-trained on {17 3D-annotated datasets}: Co3Dv2~\cite{reizenstein2021common}, BlendedMVS~\cite{yao2020blendedmvs}, DL3DV~\cite{ling2024dl3dv}, MegaDepth~\cite{li2018megadepth}, Kubric~\cite{greff2022kubric}, WildRGB~\cite{xia2024rgbd}, ScanNet~\cite{dai2017scannet}, HyperSim~\cite{roberts2021hypersim}, Mapillary~\cite{antequera2020mapillary}, Habitat~\cite{szot2021habitat}, Replica~\cite{karaev2023dynamicstereo}, MVS-Synth~\cite{huang2018deepmvs}, PointOdyssey~\cite{zheng2023pointodyssey}, Virtual KITTI~\cite{cabon2020virtual}, Aria Synthetic Environments~\cite{pan2023aria}, Aria Digital Twin~\cite{pan2023aria}, and an Objaverse~\cite{deitke2023objaverse}-like synthetic asset set, all providing ground-truth cameras, depths, and pointmaps. V-DPM then fine-tunes both the backbone and geometry/tracking heads on {6 additional 3D/4D-annotated datasets}: ScanNet++~\cite{yeshwanth2023scannet++} and BlendedMVS~\cite{yao2020blendedmvs} for static scenes, and Kubric-F~\cite{greff2022kubric}, Kubric-G~\cite{greff2022kubric}, PointOdyssey~\cite{zheng2023pointodyssey}, and Waymo~\cite{sun2020scalability} for dynamic scenes. The heads use the representations from the pre-trained VGGT~\cite{wang2025vggt} backbone and are further fine-tuned, so they ultimately benefit from {23 datasets} in total.

In contrast, \ours\ is initialized from Wan2.1-T2V~\cite{wan2025wan}, a video diffusion transformer pre-trained on billions of generic web images and videos with \emph{no 3D annotations of any kind}, and fine-tuned on only {4 synthetic 3D/4D datasets} (Kubric~\cite{greff2022kubric}, PointOdyssey~\cite{zheng2023pointodyssey}, and Dynamic Replica~\cite{karaev2023dynamicstereo} for dynamic scenes, and TartanAir~\cite{wang2020tartanair} for static scenes). \textbf{The 3D/4D supervision seen by \ours\ during training is thus a small fraction of that seen by V-DPM (23 datasets \emph{vs.} 4 datasets).}

Despite the dataset scale gap, \ours\ + V-DPM achieves competitive $\text{APD}_\text{3D}$, while exceeding it in AJ. This demonstrates that the spatio-temporal priors learned from large-scale generic video data effectively compensate for the absence of dense 3D supervision, serving as a strong foundation for 3D tracking. We attribute the small remaining gap to the dataset-scale difference. Even when we use frame-anchored reconstruction maps from V-DPM as input, we only access its 3D point predictions, not its pre-trained representations, while V-DPM's tracking head starts from pre-trained representations trained on 17 datasets. Furthermore, we anticipate that with access to stronger 3D foundation models in the future, \ours\ can achieve even better performance.

\begin{wraptable}[8]{r}{0.40\textwidth}
\centering
\vspace{-20pt}
\caption{\textbf{Inference efficiency} with different frame lengths.}
\vspace{3pt}
\label{tab:vdpm_efficiency}
\resizebox{\linewidth}{!}{
\begin{tabular}{c|lcc}
\toprule
Frames & Method & Time (s)$\downarrow$ & Memory (GB)$\downarrow$ \\
\midrule
12 & V-DPM~\cite{sucar2026v} & 12.32 & 12.60 \\
12 & \textbf{\ours}                        & \textbf{  3.91   }  & \textbf{7.63} \\
\midrule
23 & V-DPM~\cite{sucar2026v} & 51.49  & 	17.69 \\
23 & \textbf{\ours}                   & \textbf{7.84}  & \textbf{7.63} \\
\bottomrule
\end{tabular}
}
\end{wraptable}

\paragrapht{Inference Efficiency.}
\ours\ is substantially more efficient than V-DPM. As shown in Tab.~\ref{tab:vdpm_efficiency}, evaluated at $448 \times 448$ resolution on 12- and 23-frame clips using a single A6000 GPU, the efficiency gap widens as the clip length $L$ grows. For a 12-frame clip, \ours\ runs \textbf{3.2$\times$ faster} and uses \textbf{1.7$\times$ less memory}. For a 23-frame clip, the gap widens to \textbf{6.6$\times$ faster} and \textbf{2.3$\times$ less memory}.

V-DPM predicts track pointmaps $\{P_0(t_j)\}_{j=0}^{L-1}$ via an attention-based, time-conditioned decoder invoked once per timestamp $t_j$. Running the decoder $L$ times, with each call performing self-attention over all $L$ frames, incurs $\mathcal{O}(L^2)$ time and $\mathcal{O}(L)$ memory. In contrast, \ours\ predicts all trajectories in a \emph{single feed-forward} pass within the \emph{compressed latent space} of a video DiT. For longer clips, \ours\ uses interleaved inference with a fixed clip length, yielding $\mathcal{O}(L)$ runtime and $\mathcal{O}(1)$ peak memory. This efficiency gap is decisive for long-video applications.

\paragrapht{Summary.}
\ours\ trades a small amount of point accuracy for (i) data efficiency, requiring only 4 synthetic 3D/4D-annotated datasets for fine-tuning compared to V-DPM's 23 3D/4D-annotated datasets; (ii) compatibility with any 3D geometry estimator, naturally benefiting from future advances in 3D foundation models, including V-DPM itself; and (iii) substantial efficiency gains, particularly for long videos. 

\section{Additional Attention Visualization}
\label{sec:appendix_attention}

\begin{figure*}[p]
\centering
\includegraphics[width=\textwidth]{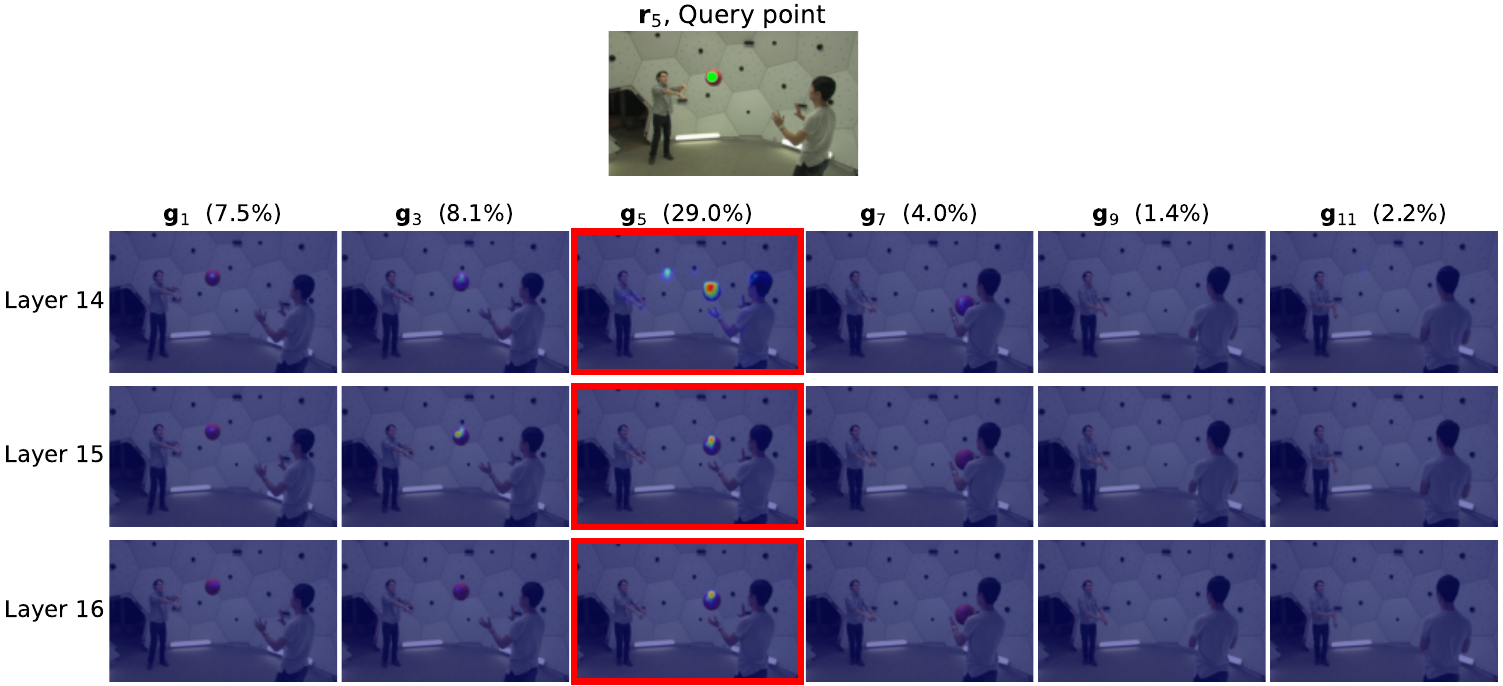}
\caption{\textbf{Query–key attention visualization on PStudio~\cite{joo2015panoptic}.} The query point is marked with a green circle on the baseball. We visualize attention from the track latent $\mathbf{r}_5$ at timestamp $t_5$ to geometry latents $\{\mathbf{g}_j\}_{j=0}^{F}$ across transformer layers. Attention is concentrated on the temporally aligned geometry latent $\mathbf{g}_5$ (29.0\%), showing that RoPE correctly assigns a target timestamp to each track latent.}
\label{fig:qualitative1}
\end{figure*}

\begin{figure*}[p]
\centering
\includegraphics[width=\textwidth]{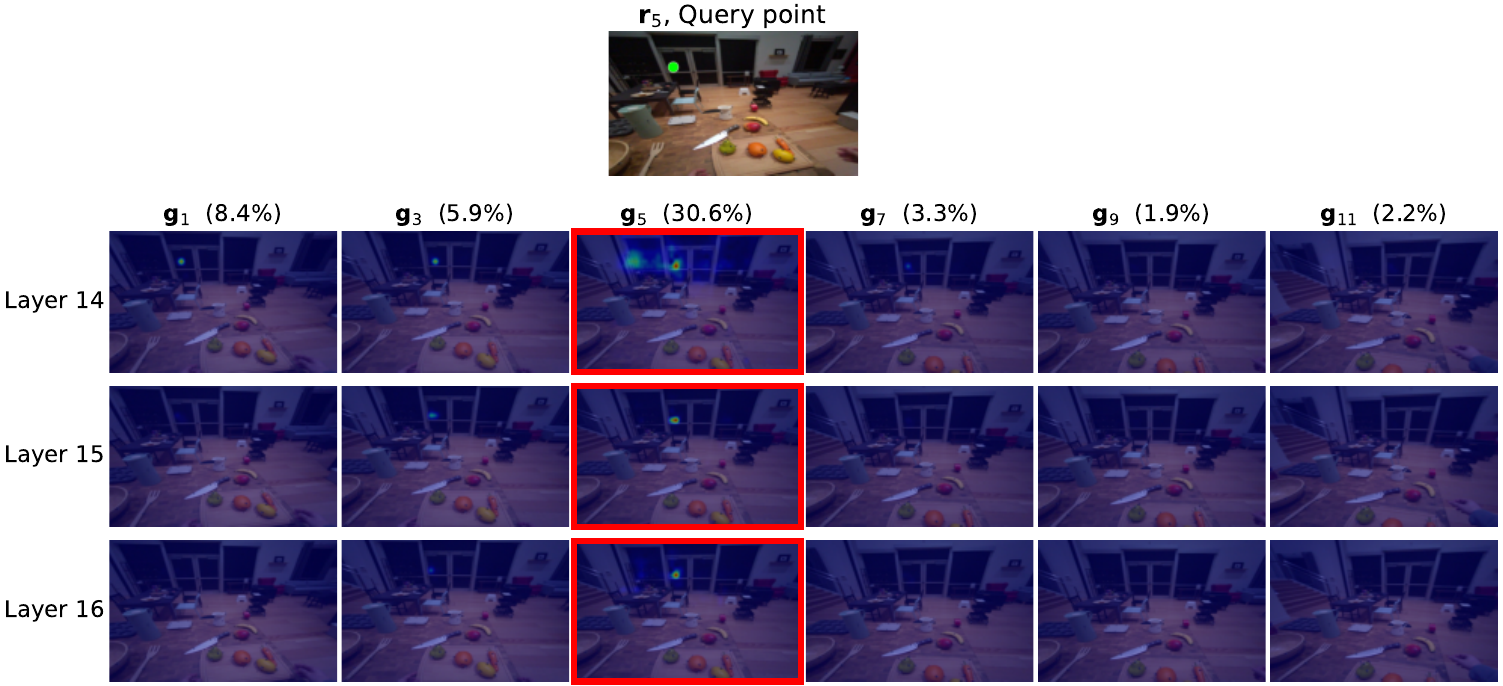}
\caption{\textbf{Query–key attention visualization on ADT~\cite{pan2023aria}.} The query point is marked with a green circle on the left door. We visualize attention from the track latent $\mathbf{r}_5$ at timestamp $t_5$ to geometry latents $\{\mathbf{g}_j\}_{j=0}^{F}$ across transformer layers. Attention is concentrated on the temporally aligned geometry latent $\mathbf{g}_5$ (30.6\%), showing that RoPE correctly assigns a target timestamp to each track latent.}s
\label{fig:qualitative2}
\end{figure*}

\paragrapht{Temporal alignment between track and geometry latents.}
Fig.~\ref{fig:qualitative1} and Fig.~\ref{fig:qualitative2} visualize the query–key attention from a track latent $\mathbf{r}_5$ to geometry latents $\{\mathbf{g}_k\}_{k=0}^{F}$ across transformer layers. The red box marks the temporally aligned geometry latent $\mathbf{g}_5$. We observe that each track latent $\mathbf{r}_i$ assigns the highest attention to its corresponding geometry latent $\mathbf{g}_i$. We quantify this by averaging attention mass over all transformer layers: the temporally aligned geometry latent receives the highest attention (29.0\% in Fig.~\ref{fig:qualitative1} and 30.1\% in Fig.~\ref{fig:qualitative2}). This verifies that temporal RoPE alignment provides a reliable signal for identifying the correct timestamp.

\begin{figure*}[p]
\centering
\includegraphics[width=\textwidth]{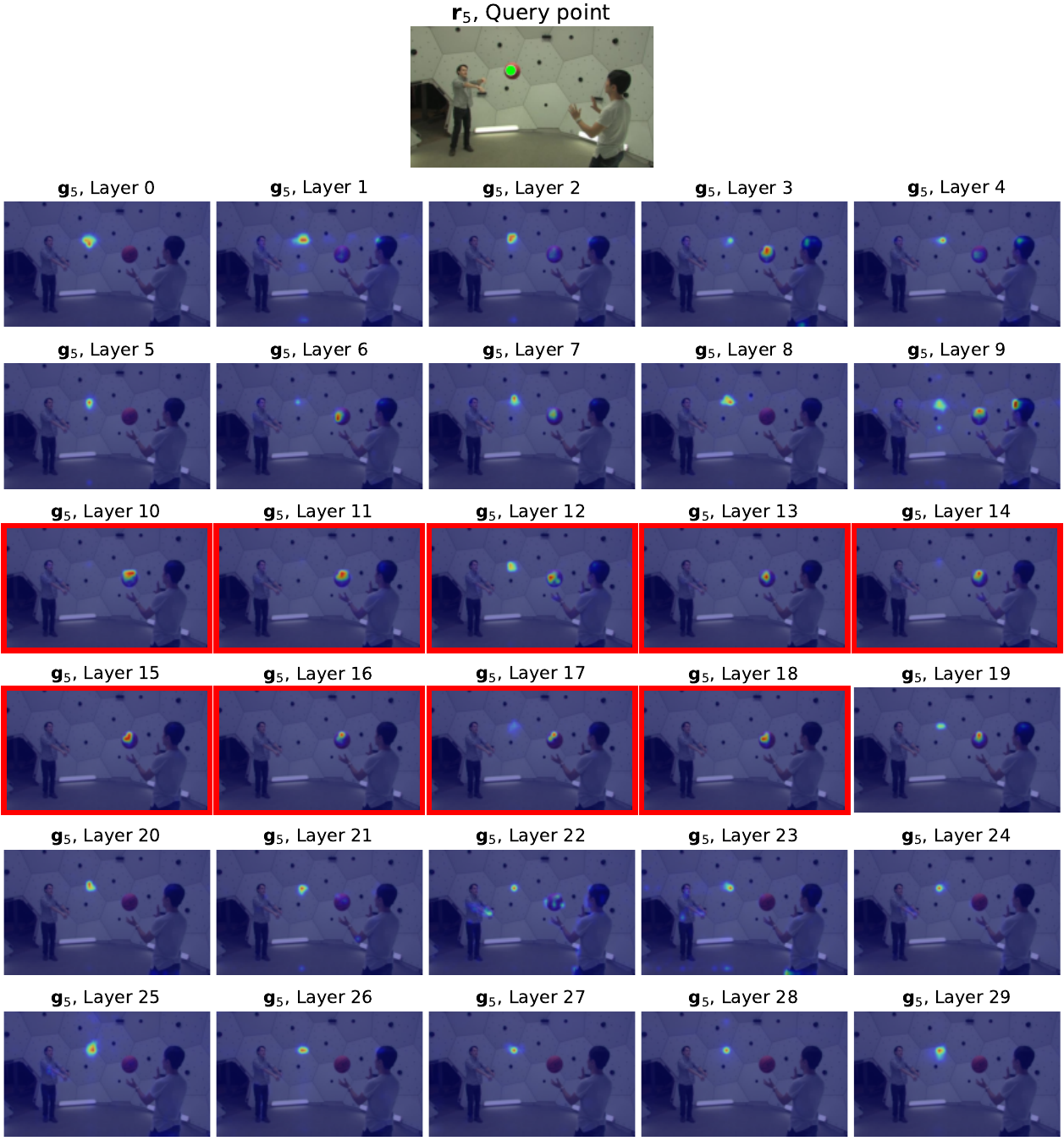}
\caption{\textbf{Query–key attention visualization on PStudio~\cite{joo2015panoptic}.} The query point is marked with a green circle. Attention between the temporally aligned track and geometry latents identifies accurate correspondences of the same physical points in specific layers (highlighted in red boxes).}
\label{fig:qualitative3}
\end{figure*}

\begin{figure*}[p]
\centering
\includegraphics[width=\textwidth]{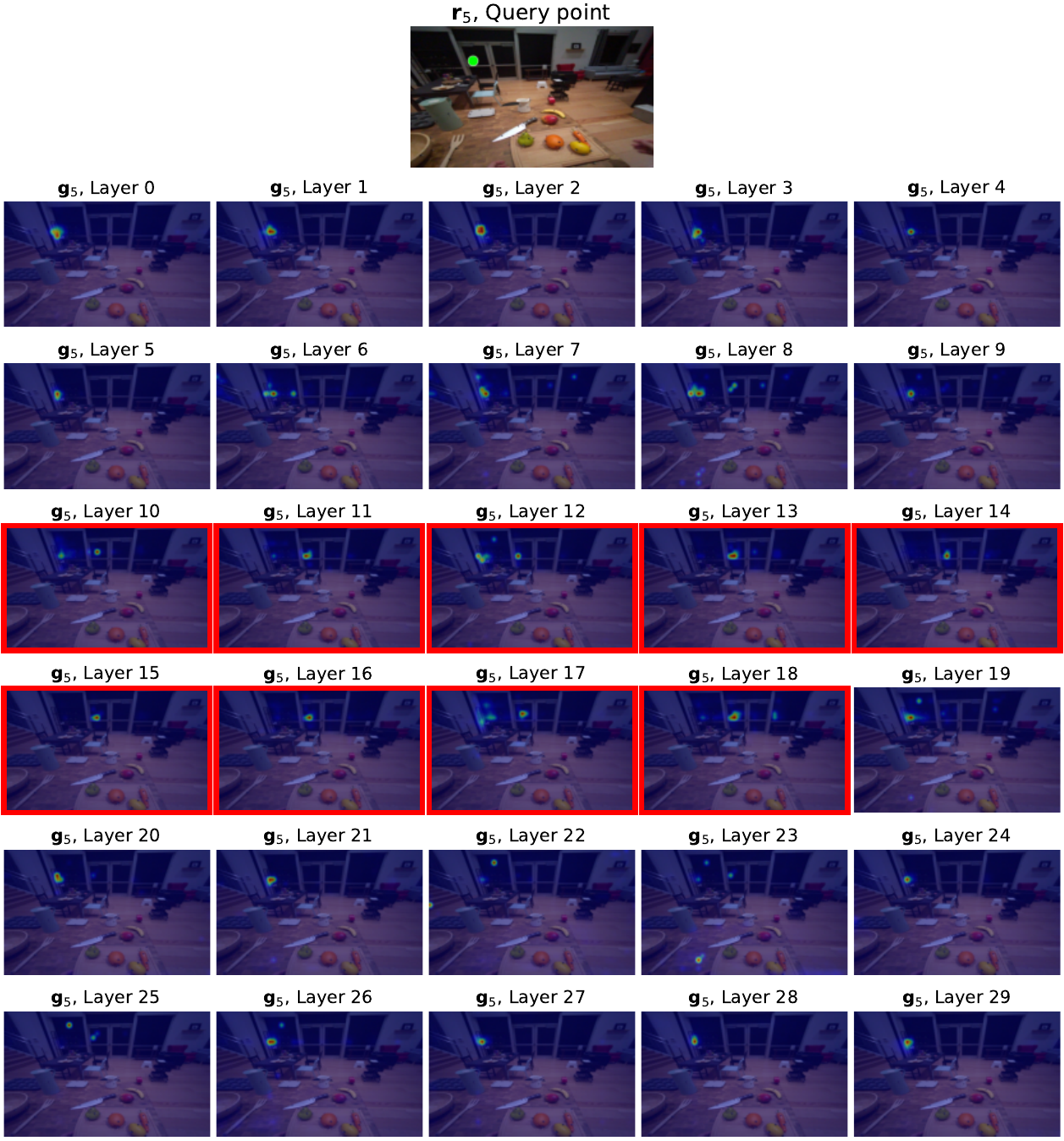}
\caption{\textbf{Query–key attention visualization on ADT~\cite{pan2023aria}.} The query point is marked with a green circle. Attention between the temporally aligned track and geometry latents identifies accurate correspondences of the same physical points in specific layers (highlighted in red boxes).}
\label{fig:qualitative4}
\end{figure*}

\paragrapht{Correspondence within aligned latents.}
Fig.~\ref{fig:qualitative3} and Fig.~\ref{fig:qualitative4} visualize attention between $\mathbf{r}_5$ and $\mathbf{g}_5$ across transformer layers. Full 3D attention establishes reliable spatial correspondences between track and geometry latents under motion (\eg, the moving baseball in Fig.~\ref{fig:qualitative3}). As discussed in prior works~\cite{nam2025emergent, son2025repurposing}, we observe layer-wise behaviors: several layers focus on RoPE-initialized positions, while a subset of layers (highlighted in red) finds correspondences between the same physical points. The same layers exhibit this behavior across different samples (Fig.~\ref{fig:qualitative4}), indicating that these layer-wise functions are consistent across inputs.

\section{Additional Qualitative Results}
\label{sec:appendix_qualitative}

\begin{figure*}[p]
\centering
\includegraphics[width=\textwidth]{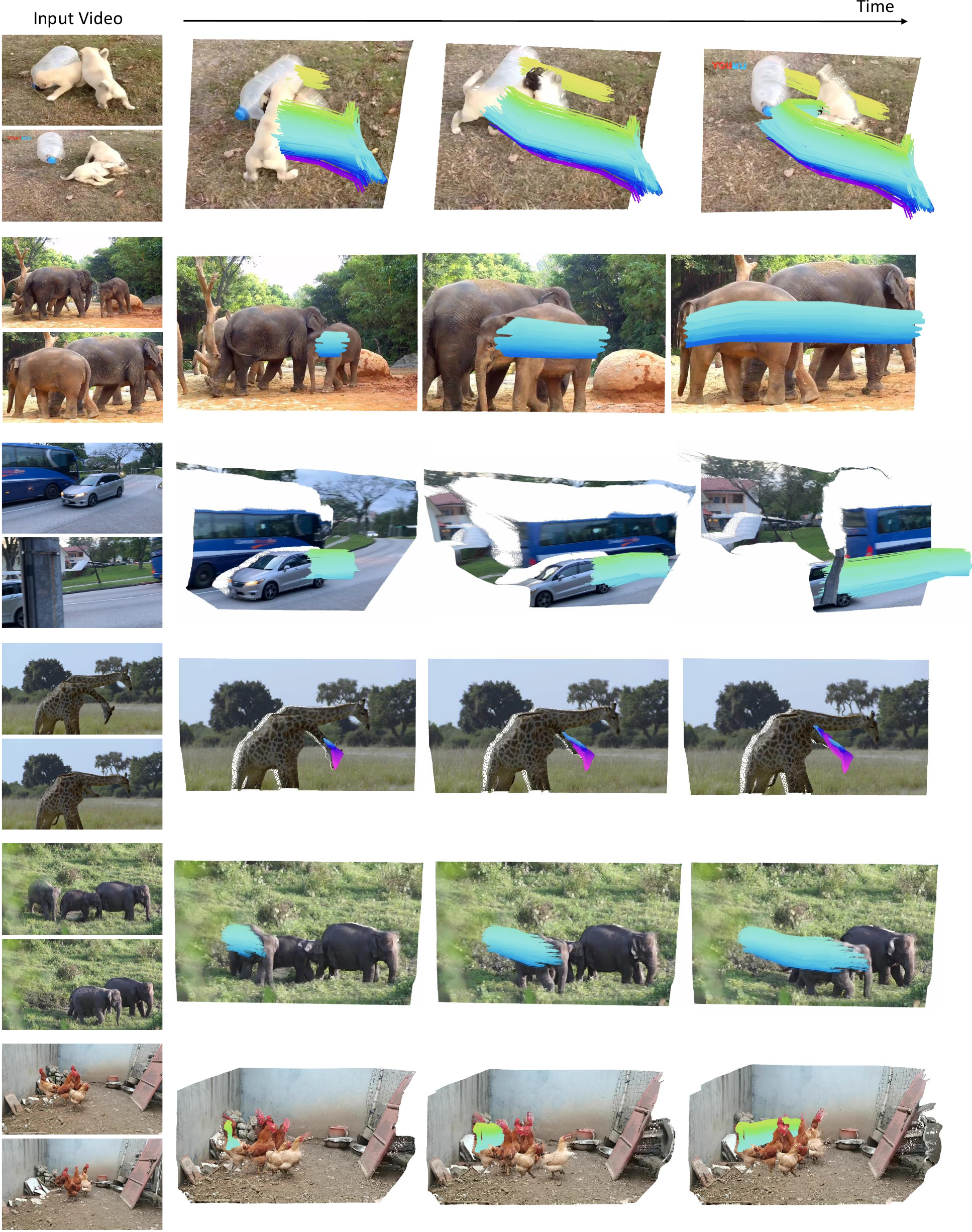}
\caption{\textbf{Qualitative results on ITTO~\cite{demler2025tracker} videos.} \ours\ accurately estimates dense 3D trajectories on real-world videos under large camera motion, object dynamics and occlusion.}
\label{fig:supp_qual_ours}
\end{figure*}

\begin{figure*}[p]
\centering
\includegraphics[width=\textwidth]{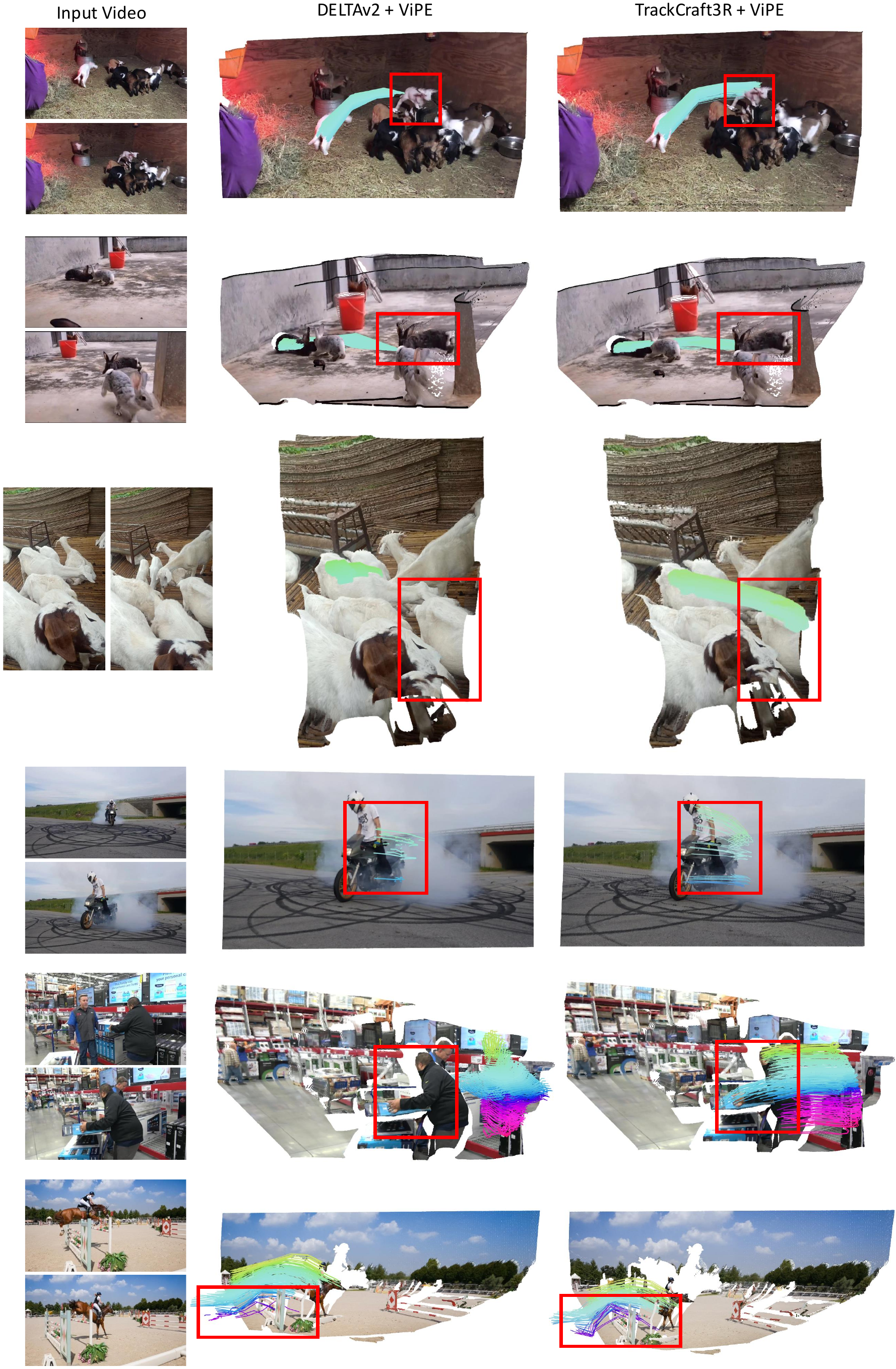}
\caption{\textbf{Qualitative comparison on ITTO~\cite{demler2025tracker} and DAVIS~\cite{pont20172017} videos.} \ours\ accurately estimates dense 3D trajectories on real-world videos under large camera motion, object motion, and occlusion. Note that the same query points are shared across methods.}
\label{fig:supp_qual}
\end{figure*}

We present additional qualitative results and comparisons with DELTAv2~\cite{ngo2025deltav2} on ITTO~\cite{demler2025tracker} and DAVIS~\cite{pont20172017} videos in Figs.~\ref{fig:supp_qual_ours} and \ref{fig:supp_qual}.

\section{Limitations and Future Work}
\label{sec:appendix_limitation}
Following the common convention in world-coordinate 3D point tracking~\cite{xiao2025spatialtrackerv2, zhang2025tapip3d, ngo2025deltav2, ngo2024delta}, \ours\ relies on per-frame depth and camera pose from external 3D foundation models~\cite{huang2025vipe, lin2025depth, li2025megasam, wang2025vggt}. While this design choice aligns with prior work, it also means that the accuracy of \ours\ is bounded by the quality of the input geometry, as shown in Tab.~\ref{tab:ablation_geometry} in the main paper. At the same time, this design allows \ours\ to benefit from future advances in 3D foundation models, as improved geometry estimators can be incorporated without retraining.

A further direction is to jointly generate video and 3D tracks, unifying 
generation and dense 4D perception within a single video DiT. Such a 
unified model could serve as a strong foundation for robotic manipulation, 
where recent work uses generated videos and tracks as intermediate 
representations for action 
prediction~\cite{huang2026pointworld, bharadhwaj2024track2act, kim2026pri4r, ko2023learning}.

Our method may have implications when applied to real-world videos involving people, such as tracking individuals. We encourage responsible use.
%%%%%%%%%%%%%%%%%%%%%%%%%%%%%%%%%%%%%%%%%%%%%%%%%%%%%%%%%%%%

\clearpage
\clearpage

\end{document}